\documentclass{article}

\usepackage{microtype}
\usepackage{graphicx}
\usepackage{subfigure}
\usepackage{booktabs} 

\usepackage{tcolorbox, soul} 
\usepackage{listings} 

\usepackage{caption}

\usepackage{hyperref}
\usepackage{enumitem}
\setlist{leftmargin=5mm}
\usepackage{xspace}

\usepackage[accepted]{icml2023}

\usepackage{amsmath}
\usepackage{amssymb}
\usepackage{mathtools}
\usepackage{amsthm}

\usepackage[capitalize,noabbrev]{cleveref}

\usepackage{amssymb,amsmath,amsthm,bbm}
\usepackage{verbatim,float,url,dsfont}
\usepackage{graphicx,subfigure,psfrag}
\usepackage{algorithm,algorithmic}
\usepackage{mathtools,enumitem}
\usepackage{multirow}
\usepackage{ragged2e}
\usepackage{array}

\usepackage[utf8]{inputenc} 
\usepackage[T1]{fontenc}    
\usepackage{booktabs}       
\usepackage{nicefrac}         
\usepackage{microtype}      
\usepackage{tikz} 
\usepackage{pgfplots}  
\pgfplotsset{compat=1.16}  

\ifdefined\TimesFont 
\usepackage{times} 
\fi

\ifdefined\ParSkip 
\usepackage{parskip} 
\fi

\theoremstyle{definition}

\newtheorem*{assumption*}{\assumptionnumber}
\providecommand{\assumptionnumber}{}


\makeatletter
\newcommand*\rel@kern[1]{\kern#1\dimexpr\macc@kerna}
\newcommand*\widebar[1]{%
  \begingroup
  \def\mathaccent##1##2{%
    \rel@kern{0.8}%
    \overline{\rel@kern{-0.8}\macc@nucleus\rel@kern{0.2}}%
    \rel@kern{-0.2}%
  }%
  \macc@depth\@ne
  \let\math@bgroup\@empty \let\math@egroup\macc@set@skewchar
  \mathsurround\z@ \frozen@everymath{\mathgroup\macc@group\relax}%
  \macc@set@skewchar\relax
  \let\mathaccentV\macc@nested@a
  \macc@nested@a\relax111{#1}%
  \endgroup
}
\makeatother

\DeclareMathOperator*{\argmin}{argmin}

\DeclareMathOperator{\rank}{rank}

\def\N{\mathbb{N}}
\def\R{\mathbb{R}}

\def\E{\mathbb{E}}
\def\P{\mathbb{P}}

\def\cA{\mathcal{A}}
\def\cB{\mathcal{B}}
\def\cC{\mathcal{C}}

\def\cG{\mathcal{G}}

\def\cH{\mathcal{H}}

\def\cN{\mathcal{N}}

\def\ind#1{\mathds{1}\left\{#1\right\}}

\newcommand{\showcomments}{yes}
\newcommand{\system}{Chatbot Arena\xspace}

\newcommand\todo[1]{\ifthenelse{\equal{\showcomments}{yes}}{{\color{red} TODO: #1}}{\ignorespaces}}
\newcommand\joey[1]{\ifthenelse{\equal{\showcomments}{yes}}{{\color{blue} (Joey: #1)}}{\ignorespaces}}
\newcommand\lianmin[1]{\ifthenelse{\equal{\showcomments}{yes}}{{\color{blue} Lianmin: #1}}{\ignorespaces}}
\newcommand\ana[1]{\ifthenelse{\equal{\showcomments}{yes}}{{\color{red} Anastasios: #1}}{\ignorespaces}}
\newcommand\ion[1]{\ifthenelse{\equal{\showcomments}{yes}}{{\color{blue} Ion: #1}}{\ignorespaces}}
\newcommand\weilin[1]{\ifthenelse{\equal{\showcomments}{yes}}{{\color{cyan} Wei-Lin: #1}}{\ignorespaces}}
\newcommand\ying[1]{\ifthenelse{\equal{\showcomments}{yes}}{{\color{purple} Ying: #1}}{\ignorespaces}}
\newcommand\tianle[1]{\ifthenelse{\equal{\showcomments}{yes}}{{\color{orange} Tianle: #1}}{\ignorespaces}}
\newcommand\dl[1]{\ifthenelse{\equal{\showcomments}{yes}}{{\color{orange} DL: #1}}{\ignorespaces}}

\icmltitlerunning{\system: An Open Platform for Evaluating LLMs by Human Preference}

\begin{document}

\twocolumn[
\icmltitle{\system: An Open Platform for Evaluating LLMs by Human Preference}



\icmlsetsymbol{equal}{*}

\begin{icmlauthorlist}
\icmlauthor{Wei-Lin Chiang}{equal,ucb}
\icmlauthor{Lianmin Zheng}{equal,ucb}
\icmlauthor{Ying Sheng}{stanford}
\icmlauthor{Anastasios N. Angelopoulos}{ucb}
\icmlauthor{Tianle Li}{ucb}
\icmlauthor{Dacheng Li}{ucb}
\icmlauthor{Banghua Zhu}{ucb}
\icmlauthor{Hao Zhang}{ucsd}
\icmlauthor{Michael I. Jordan}{ucb}
\icmlauthor{Joseph E. Gonzalez}{ucb}
\icmlauthor{Ion Stoica}{ucb}
\end{icmlauthorlist}

\icmlaffiliation{ucb}{UC Berkeley}
\icmlaffiliation{stanford}{Stanford}
\icmlaffiliation{ucsd}{UCSD}


\icmlcorrespondingauthor{Wei-Lin Chiang}{weichiang@berkeley.edu}

\icmlkeywords{Machine Learning, ICML}

\vskip 0.3in
]



\printAffiliationsAndNotice{\icmlEqualContribution}

\begin{abstract}
Large Language Models (LLMs) have unlocked new capabilities and applications; however, evaluating the alignment with human preferences still poses significant challenges.
To address this issue, we introduce \system, an open platform for evaluating LLMs based on human preferences. Our methodology employs a pairwise comparison approach and leverages input from a diverse user base through crowdsourcing. The platform has been operational for several months, amassing over 240K votes. This paper describes the platform, analyzes the data we have collected so far, and explains the tried-and-true statistical methods we are using for efficient and accurate evaluation and ranking of models. We confirm that the crowdsourced questions are sufficiently diverse and discriminating and that the crowdsourced human votes are in good agreement with those of expert raters. These analyses collectively establish a robust foundation for the credibility of \system.
Because of its unique value and openness, \system has emerged as one of the most referenced LLM leaderboards, widely cited by leading LLM developers and companies. Our demo is publicly available at \url{https://chat.lmsys.org}.
\end{abstract}

\section{Introduction}

Recent advancements in large language models (LLMs) have significantly expanded their capabilities beyond traditional natural language processing boundaries, addressing a broad array of general tasks~\cite{openai2023gpt4,team2023gemini,touvron2023llama2}. These developments underscore the potential of LLMs but also have raised concerns with respect to performance evaluation. Current benchmarks often fail to capture the nuanced and diverse aspects of these models, particularly in assessing their alignment with human preferences in real-world, open-ended tasks.

\begin{figure}
    \centering
    \includegraphics[width=1.01\columnwidth]{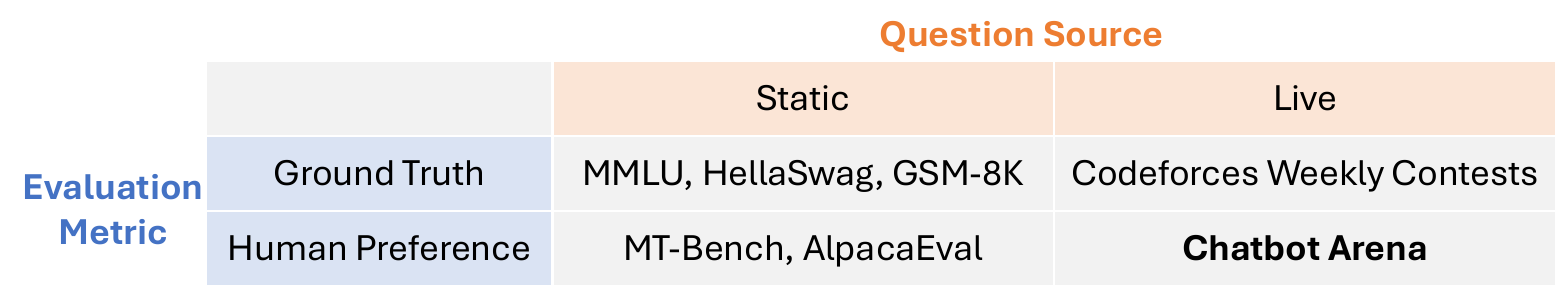}
    \vspace{-2em}
    \caption{Classification of LLM benchmarks: We categorize along two dimensions: whether the questions are from a static dataset or a live, fresh source, and whether the evaluation metric relies on ground truth or (approximated) human preferences. MMLU~\cite{hendrycks2020measuring}, HellaSwag~\cite{zellers2019hellaswag}, GSM-8K~\cite{cobbe2021gsm}, MT-Bench~\cite{zheng2023judging}, and AlpacaEval~\cite{alpaca_eval} are common examples of static benchmarks. \system is the platform introduced in this paper.}
    \vspace{-1em}
    \label{fig:eval_classification}
\end{figure}

To assess the performance of LLMs, the research community has introduced a variety of benchmarks. These benchmarks can be categorized based on two factors: the source of questions (either static or live) and the evaluation metric (either ground truth or human preference). According to these factors, benchmarks can be classified into four categories, as shown in \autoref{fig:eval_classification}. While a range of benchmarks is beneficial, the most prevalent current method for evaluating LLMs remains a static, ground-truth-based evaluation, partly because such evaluations are inexpensive and reproducible.

However, these static, ground-truth-based benchmarks exhibit several limitations. Firstly, the questions within these benchmarks are not open-ended, hindering the ability to capture the flexible and interactive use found in real-world settings~\cite{zheng2023judging}. Secondly, the test sets in these benchmarks are static, meaning they can become contaminated over time, which undermines the reliability of the evaluation results~\cite{yang2023rethinking}. Furthermore, for many complex tasks, establishing a definitive ground truth is not only challenging but sometimes unattainable. Consequently, current benchmarks fail to adequately address the needs of state-of-the-art LLMs, particularly in evaluating user preferences. Thus, there is an urgent necessity for an open, live evaluation platform based on human preference that can more accurately mirror real-world usage.

Creating such a benchmark platform entails significant challenges. It requires the collection of live, fresh, and diverse user questions to accurately represent real-world scenarios. Additionally, developing scalable, incremental, and efficient ranking systems is essential for evaluating a large number of models. Moreover, ensuring the quality of human evaluations is crucial given the noisy nature of human preferences.

To this end, we introduce \system, a benchmarking platform for LLMs that features anonymous, randomized battles in a crowdsourced setting. \system is a free website open to all users.\footnote{\url{https://chat.lmsys.org}} On this website, a user can ask a question and get answers from two anonymous LLMs.
Afterward, the user casts a vote for the model that delivers the preferred response, with the models' identities revealed only after voting. This crowdsourced method effectively gathers a diverse array of fresh user prompts, accurately reflecting real-world LLM applications. 
Armed with this data, we employ a suite of powerful statistical techniques, ranging from the statistical model of~\citet{bradley1952rank} to the E-values of~\citet{vovk2021values}, to estimate the ranking over models as reliably and sample-efficiently as possible.
With these tools in hand, we have designed efficient sampling algorithms specifically to select model pairs in a way that accelerates the convergence of rankings while retaining statistical validity.

We conduct a thorough analysis of the collected data to ensure the credibility of our platform. We demonstrate that the user-generated questions are sufficiently diverse to encompass a wide range of LLM use cases and are sufficiently challenging to differentiate between models. Furthermore, we confirm that the crowd-sourced votes are highly consistent with expert evaluations.

We have been running our system since Apr 2023 and have received over 240K votes from about 90K users in over 100 different languages as of Jan 2024.
To encourage user engagement, we have made over 50 state-of-the-art models available for free.
We also collaborate with leading model developers such as OpenAI, Google, Anthropic, Mistral, Hugging Face, and various universities, incorporating their latest models into our platform. We keep the community engaged by routinely updating the leaderboard, publishing analytical blogs, releasing datasets, and sharing information via tweets. 
Because of its unique and significant value, our leaderboard has emerged as one of the most referenced in the LLM field and has become a benchmark for the industry.
We commit to making our data and code available, ensuring that this platform is open-source and open-accessible.


We make the following contributions:
\begin{itemize}[itemsep=0.1em]
    \item We build the first large-scale crowd-sourced live LLM evaluation platform with over 1M users visit.\footnote{The number was estimated by Google Analytics as of March 2024. Note that user visit may not convert to votes as our website also offers ``direct chat'' mode.}
    \item We conduct an in-depth analysis of the collected data, including prompt diversity, quality, vote quality, and insights on human feedback.
    \item We will publicly release a human preference dataset with over 100K pairwise votes collected from \system.
    \item We design an efficient sampling algorithm that actively chooses which model pairs to show, such that our sample efficiency improves, sometimes to a large degree.
\end{itemize}

\section{Related Work}

\textbf{LLM Benchmarks.}
We briefly review the common LLM benchmarks, following the classification presented in \autoref{fig:eval_classification}. The most prevalent benchmarks are static, ground-truth-based ones, typically in the form of multiple-choice questions or question-answering tasks with predefined answers and test cases.
These benchmarks encompass a range of topics including language understanding, mathematics, coding, and logical reasoning. Prominent examples in this category are MMLU~\cite{hendrycks2020measuring}, HellaSwag~\cite{zellers2019hellaswag}, GSM-8K~\cite{cobbe2021gsm}, BigBench~\cite{srivastava2023beyond}, AGIEval~\cite{zhong2023agieval}, and HumanEval~\cite{chen2021evaluating}. Benchmarks focusing on safety, such as ToxicChat~\cite{toxicchat}, and comprehensive suites like HELM~\cite{liang2022holistic}, also exist.
In addition to closed-ended questions, benchmarks can include open-ended questions that are evaluated by human judgment, which can be rated by experts or crowd workers such as Amazon Mechanical Turk~\cite{karpinska-etal-2021-perils,koala_blogpost_2023,wang-etal-2023-self-instruct}.
The recent trend includes utilizing GPT-4 for approximating human judgment~\cite{chiang-lee-2023-large}, with notable instances being MT-Bench~\cite{zheng2023judging} and AlpacaEval~\cite{alpaca_eval}.
In addition to static benchmarks, live benchmarks that include fresh questions are also available. These questions can be obtained from annual exams or weekly online contests such as Codeforces~\cite{li2022competition, huang2023competition}. They can also be sourced from human interaction. Some studies have explored using live human interaction for reinforcement learning from human preference~\cite{bai2022training,ouyang2022training,touvron2023llama2}. However, these studies are typically limited to specific organizations.
In this paper, we introduce \system, the first open, large-scale, and crowdsourced benchmark platform that utilizes live human interaction.

\textbf{Risks of Static Benchmarks.}
Static benchmarks have certain issues, including contamination, saturation, overfitting, and a lack of human alignment~\cite{yang2023rethinking,oren2023proving}. DynaBench~\cite{kiela2021dynabench} identifies these challenges and recommends the use of a live benchmark that incorporates a human-in-the-loop approach for classical NLP benchmarks. Our system adopts a similar spirit. However, our focus is on chatting with LLMs, and we implement this on a significantly larger user scale.

\textbf{Ranking System.}
Ranking systems have been a well-studied topic in statistics. Related topics include probability models~\cite{mm_bradley_terry,tie_in_bradley_terry}, rank elicitation~\cite{online_rank,pmlr-v32-busa-fekete14,preference_rank_elicitation}, and online experiment design~\cite{sequential_design, karimi2021online}.
The Elo rating system has also been used for LLMs~\cite{bai2022training,boubdir2023elo}.
Contributing to this literature, we introduce techniques for accelerating ranking convergence and detecting abnormalities, specifically applied to large-scale, real-world settings of LLMs.

\textbf{Human Preference Dataset.}
Owing to the significance of human preferences, several datasets and analyses exist that incorporate human preferences. These include OpenAssistant~\cite{kopf2023openassistant}, HH-RLHF~\cite{bai2022training}, LMSYS-Chat-1M~\cite{zheng2023lmsyschat1m}, and synthetic approximations of human preferences like UltraFeedback~\cite{cui2023ultrafeedback} and Nectar~\cite{starling2023}.
Our prior data release, LMSYS-Chat-1M~\cite{zheng2023lmsyschat1m}, is similarly collected via crowdsourcing. However, LMSYS-Chat-1M comprises solely conversations and lacks human preference data, rendering it unsuitable for direct use in ranking studies. This paper focuses on the analysis of preference data for ranking purposes.

\section{Human Preference Data Collection}

In this section, we discuss our interface design to collect human preferences and present summary statistics.
\subsection{Interface}

\system crowd-sources feedback from users for model evaluation. Our goal is to design an ease-of-use interface to reduce friction for users to contribute data.
Since we collect feedback from many users, it is difficult to set a consistent grading rubric across different people. Hence, we adopt a pairwise comparison mechanism where users only need to compare two model responses and vote for the better one, instead of requiring users to provide an absolute score.

In each battle, two anonymous models are sampled. To encourage data diversity, we do not preset any input prompt on the website. Users are free to input any prompt to the two models. We believe this creates incentives for user engagement, particularly given that we offer a free service. It also helps us collect a diverse set of inputs representing real-world usage. After models provide their answers, user compare them side-by-side and vote for the preferred answer. If a user cannot choose in the first turn, the user can continue chatting until identifying a winner. For those who are unsure, we also present two buttons, ``tie'' or ``both are bad.'' Figure~\ref{fig:screenshot-arena} shows a screenshot of our interface.
Before using our service, users are required to accept terms of use, which gives us their consent to release the data publicly.

\subsection{Data Statistics}

We began collecting data in April 2023. As of Jan 2024, we have received around 240K votes from over 90K users. Our data involves more than 50 models, including both proprietary models like GPT-4, Claude, and Gemini, as well as open models such as LLaMA and Mistral.
These conversations cover more than 100 languages, with 77\% being in English, 5\% in Chinese, and the remaining languages, such as Russian, German, Spanish, French, and Japanese, each representing less than 2\% of the total.
Each data point includes multi-turn conversations between the user and two LLMs, and a vote to indicate which model the user prefers.
We summarize statistics in Table~\ref{table:data_stats} along with other existing human preference datasets.



\begin{figure*}
\centering
\vspace{-5px}
\includegraphics[width=0.48\textwidth]{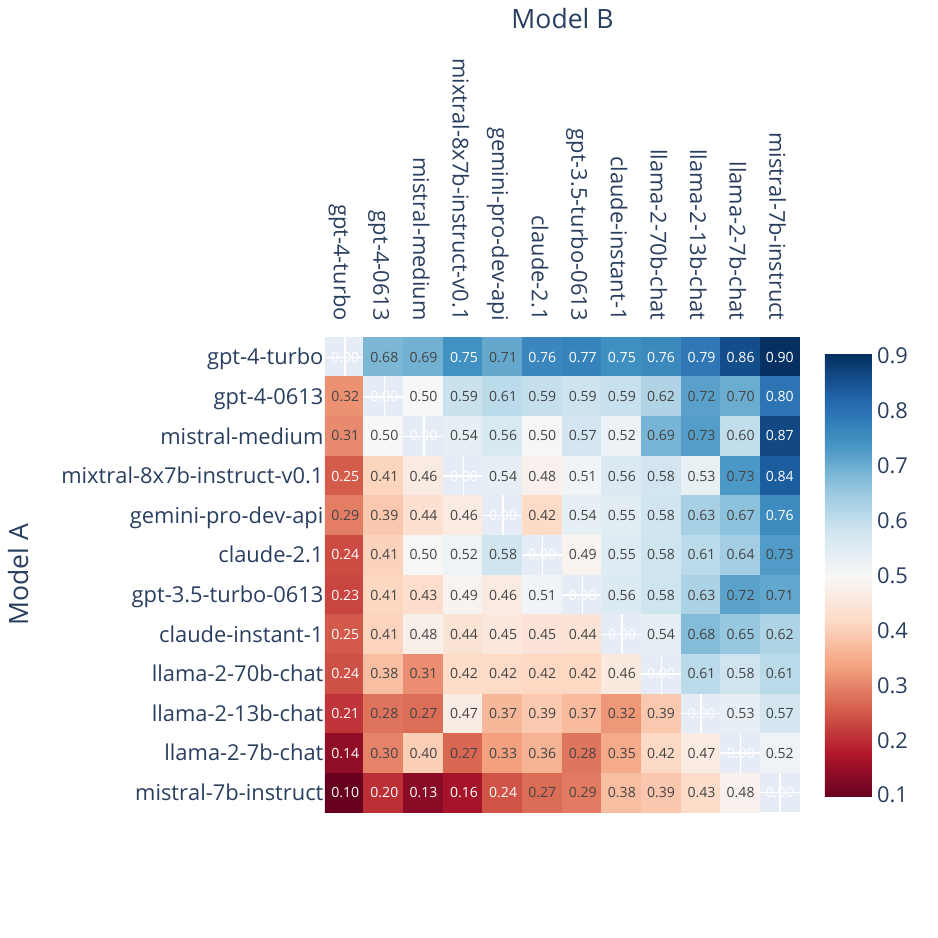}
\hfill
\includegraphics[width=0.48\textwidth]{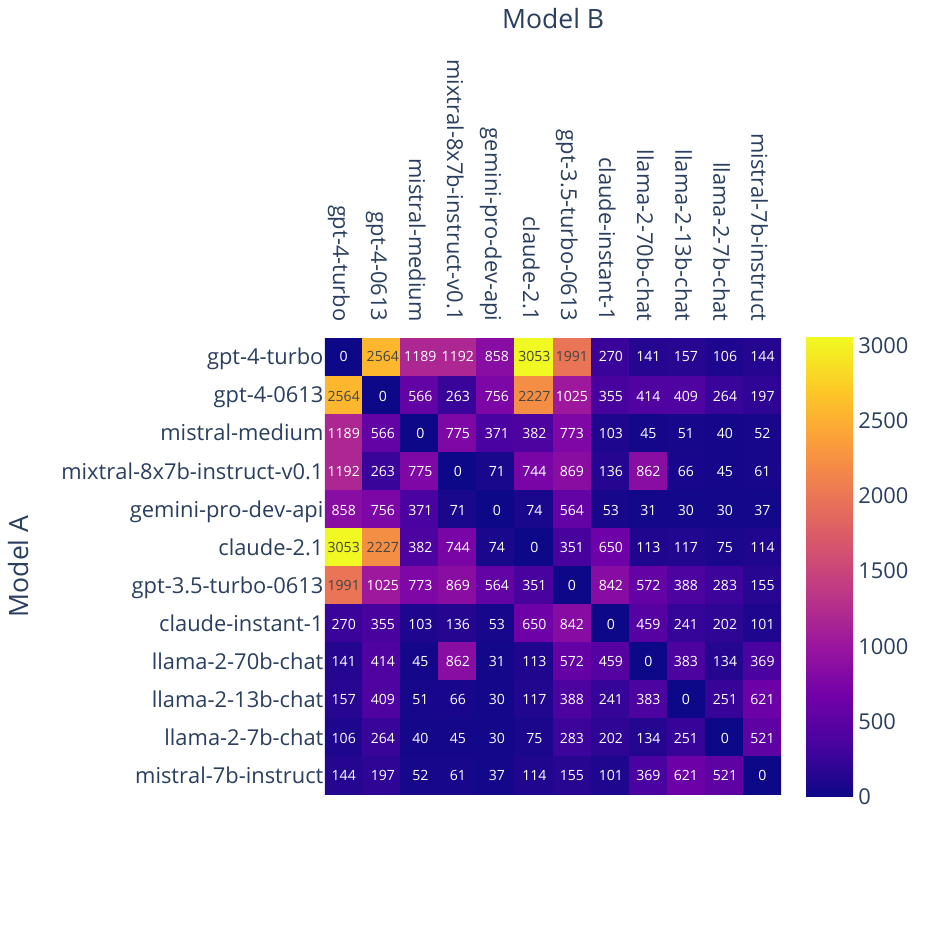}
\vspace{-40px}
\caption{Win-rate (left) and battle count (right) between a subset of models in \system.}
\label{fig:winrate-match-count}
\end{figure*}

Figure~\ref{fig:vote-cnt} in the Appendix shows the vote count per model. On average, 8K votes are collected for each model.
In Figure~\ref{fig:winrate-match-count}, we select a set of representative models and present their win rate and the number of battles. Note that we employ non-uniform sampling to concentrate votes on model pairs that have similar performance due to higher uncertainty. 
This helps us reduce the number of votes required to reach stable results. 
We later develop an adaptive sampling method and demonstrate its effectiveness against random sampling. See Section~\ref{sec:efficient-estimation} for further analysis.


To ensure anonymity, we use keywords to filter out conversations containing model identity such as model name (e.g., GPT, Claude) or companies (e.g., OpenAI, Anthropic).
To avoid misuse, we adopt OpenAI moderation API to flag conversations that contain unsafe content.
The flagged user requests account for 3\% of the total requests.
Figure~\ref{fig:daily-vote} in the Appendix shows the number of valid user votes over time, where we get 1-2K votes per day in recent months and spikes as we introduce new models or leaderboard updates.




\begin{table*}[t]
\centering
\vspace{-5px}
\caption{Statistics of human preference datasets, including Anthropic HH~\citep{bai2022training}, OpenAssistant Conversations~\citep{kopf2023openassistant}, and \system (as of 2024/1/21). The tokens are counted by Llama2's tokenizer. ``Conv'' = Conversation. ``Lang'' = Language.}
\label{table:data_stats}
\footnotesize
\begin{tabular}{lrrrrrrrr}
\toprule
\multirow{2}{*}{Dataset} & \multirow{2}{*}{\# Convs} & \multirow{2}{*}{\# Models} &  \multirow{2}{*}{\# Users} & \multirow{2}{*}{\# Langs} & Avg. \# Turns & Avg. \# Tokens & Avg. \# Tokens\\
&  &  & &  & per Sample & per Prompt & per Response\\
\midrule
Anthropic HH   & 338,704   & -  & 143     & 1   & 2.3 & 18.9 & 78.9  \\
OpenAssistant  & 66,497    & -  & 13,500  & 35  & -   & 36.9 & 214.2 \\
\midrule
\system (20240121)  & 243,329 & 50 & 90,051 & 149 & 1.3 & 94.9  & 269.0 \\
\bottomrule
\end{tabular}
\end{table*}

\section{From Pairwise Comparisons to Rankings}
\label{sec:rankings}

Our data consists of pairwise comparisons---but how can we use these comparisons to recover a ranking over all $M$ models?
This is a well-studied topic in the literature on learning to rank~\cite{liu2009learning}, and we present our perspective here.
We let $\cA = \{ (m, m') : m < m' \text{ and } m,m' \in [M] \}$ denote our comparative data set.

We consider a sequential setting, where at time $t \in \N$, we serve the human a pair of models $A_t \in \cA$ (which we pick), and in turn we observe the human's response $H_t \in [0,1]$.
As an example, we might have that $A_t = (1,2)$ and $H_t = 1$, indicating that the human prefers model 2 over model 1.
In the ensuing text, we will primarily focus on the binary case---where $H_t \in \{0, 1\}$---but our approach will generalize to any form of feedback, including the possibility of allowing the human to express different degrees of preference or to say the models are tied. 

One critical goal is to estimate the \emph{win matrix}: $\theta^*(a) = \E[H_t \mid A_t = a]$, for all $a \in \cA$; see the left panel of Figure~\ref{fig:winrate-match-count} for an illustration of the (empirical) win matrix.
In the binary case, the $a$ entry in the win matrix corresponds to the probability the human prefers model $a_2$ to $a_1$ when shown the pair $a$.
Finding the win matrix is a relatively straightforward mean-estimation problem; we will provide details in Section~\ref{sec:efficient-estimation}.

Formally, consider a \emph{score} $s(\P) \in \R^M$, where $\P$ is a joint distribution over $A$ and $H$ (by default, we will target a uniform distribution over $\cA$).
Each model has a true score $s(\P)_m$, and better models will have higher scores.
In particular, we have the rank of model $m$: 
\begin{equation}
    \rank(\P)_m = 1+\sum\limits_{m' \in [M]} \ind{s(\P)_{m'} > s(\P)_m}.
\end{equation}
The best model has rank $1$. 
If there is another model tied for best, they will both get assigned rank $1$.

\textbf{Picking a score.}
A standard score function in this setting is the vector of Bradley-Terry (BT) coefficients~\cite{bradley1952rank}.
In the Bradley-Terry model, $H_t \in \{0,1\}$, and the probability model $m$ beats model $m'$ is modeled via a logistic relationship:
\begin{equation}
    \P(H_t = 1) = \frac{1}{1+e^{\xi_{m'}-\xi_m}},
\end{equation}
where $\xi$ is an $M$-length vector of so-called BT coefficients.
Without loss of generality, we take $\xi_1 = 0$ (since the model is invariant to addition in $\xi$).
Our goal is to estimate the population Bradley-Terry coefficients, i.e., those that minimize the binary cross-entropy:
\begin{equation}
    \label{eq:population-BT}
    s(\P) = \argmin_{\xi} \E_{(A,H) \sim \P}\left[\ell\left(H, \frac{1}{1+e^{\xi_{A_2}-\xi_{A_1}}}\right)\right],
\end{equation}
where $\ell$ is the binary cross-entropy loss, $\ell(h,p) = -(h\log(p) + (1-h)\log(1-p))$.

Although the BT model technically assumes a parametric form for the model win rates, the seminal results of~\citet{huber1967behavior, white1982maximum} show that maximum likelihood estimators are still asymptotically normal even when these assumptions \emph{do not} hold, so long as the so-called ``sandwich'' covariance matrix is used; see Section~\ref{sec:efficient-estimation} for details, and see Appendix~\ref{app:np-bt} for a nonparametric extension of the Bradley-Terry model.
Finally, we remark that previous evolutions of our online interface have reported different ranking scores, such as the Elo score~\cite{elo1967proposed} instead of the BT coefficients.
We made this change because the BT coefficients are better for the purpose of statistical estimation.

\section{Efficient Approximate Ranking}
\label{sec:efficient-estimation}

In Section~\ref{sec:rankings} we described how to calculate the win matrix, score, and rank.
Now we describe our estimation procedures.

\textbf{Win matrix estimation.} Estimation of the win matrix is relatively straightforward. Define $X_t(a) = \frac{1}{P_t(a)} H_t \ind{A_t = a}$, where $P_t(a)$ is the probability of sampling pair $a$ at time $t$, and $X_t$ as the according vector.
Then the estimator is
\begin{equation}
    \label{eq:reweighted-standard-estimator}
    \hat{\theta}_T = \frac{1}{T}\sum\limits_{t=1}^TX_t.
\end{equation}
Note that $\E[X_t(a)] = \theta^*(a)$ for all $t$, and thus $\hat{\theta}_T$ is an unbiased estimator of $\theta^*$.
We will furthermore estimate the covariance matrix as
\begin{equation}
    \label{eq:reweighted-covariance_matrix}
    \widehat{\Sigma}_T = \frac{1}{T}\sum\limits_{t=1}^T (X_t - \hat{\theta}_T)(X_t - \hat{\theta}_T)^\top.
\end{equation}
Under the appropriate regularity conditions, we have that
\begin{equation}
    \sqrt{T}\widehat{\Sigma}^{-1/2}(\hat{\theta} - \theta^*) \to \cN(0, I_d),
\end{equation}
and we construct confidence intervals accordingly.
For an understanding of the appropriate regularity conditions, see~\citet{durrett2019probability}, Theorem 8.2.8, where condition (ii) is trivially satisfied so long as $P_t(a) > \epsilon > 0$, and condition (i) is implied by the almost-sure convergence of $P_t(a)$ to a limiting distribution $P(a)$.

\textbf{Estimating the BT scores.}
To estimate the BT coefficients, mirroring~\eqref{eq:population-BT}, we perform (reweighted) maximum likelihood estimation on our data points:
\begin{equation}
    \label{eq:empirical-BT}
    s(\hat{\P}) = \argmin_{\xi} \sum\limits_{t=1}^T \frac{1}{P(A_t)}\ell\left(H_t, \frac{1}{1+e^{\xi_{A_{t,2}}-\xi_{A_{t,1}}}}\right),
\end{equation}
where $A_t \sim P$.
We perform the inverse weighting by $P(A_t)$ because this allows us to target a score with a uniform distribution over $A$.

To compute confidence intervals on the BT coefficients, we employ two strategies: (1) the pivot bootstrap~\cite{diciccio1996bootstrap}, and (2) the ``sandwich'' robust standard errors outlined in~\citet{huber1967behavior} (see also~\citet{freedman2006so} for an outline of the necessary technical assumptions).
Ultimately, based on the results of a simulation study described in Appendix~\ref{app:simulation-study}, we choose to deploy the sandwich intervals due to their smaller size in large samples.

\textbf{Approximate rankings.}
Finally, we report an approximate ranking for each model that accounts for the uncertainty in the estimation of the score.
Given an $M$-dimensional confidence set $\cC$ satisfying
\begin{equation}
    \P( s(\P) \in \cC ) \geq 1-\alpha,
\end{equation}
we extract an approximate ranking $R_m = 1 + \sum_{m' \in [M]} \ind{\inf \cC_{m'} > \sup \cC_{m}}$.
The uniform validity of $\cC$ directly implies that $\P( \exists m : R_m > \rank(\P)_m ) \leq \alpha$---i.e., with high probability, no model's performance is understated.
A guarantee on the other side---that no model's performance is overstated---is possible by interchanging the $\inf$ and $\sup$.
To get the uniform confidence set, we construct the chi-squared interval implied by the central limit theorem using the sandwich estimate of the variance.
In other words, we construct the interval $\{ \xi : T\left\|\hat{V}^{-1/2}(\hat\xi - \xi)\right\| \leq \chi^2_{1-\alpha, M-1}$,
where $\hat\xi$ is our MLE of the BT coefficients and $\hat{V}_{\xi}$ is the sandwich variance of the logistic regression. 

\textbf{Active sampling rule.} Our sampling rule was to choose the model pair $a\in\cA$ proportionally to the reduction in confidence interval size by sampling that pair:
\begin{equation}
    P_t(a) \propto \sqrt{\frac{\hat{\Sigma}_{t, a,a}}{|\{ t : A_t = a\}|}} - \sqrt{\frac{\hat{\Sigma}_{t, a,a}}{|\{ t : A_t = a\}| + 1}}.
\label{eq:adaptive}
\end{equation}

\subsection{Detecting Anomalous Users}
\label{sec:outlier}
On a different note, we take a first step towards identifying anomalous IP addresses in our dataset.
In a dataset of $U$ unique IPs, we let $\mathsf{IP} = \{1, \ldots, U\}$ be the set of all IP addresses.
Consider a ``test'' user, outside this database, who gives ratings $H'_1, \ldots, H'_n$ when presented actions $A'_1, \ldots, A'_n$.
The idea of our procedure is to compare the distribution of ratings for the new user to the historical distribution of ratings for a given action.
We let $\cH_a = \{ H_t : A_t = a \}$ and every time a user submits a vote, we calculate the following number:
\begin{equation}
    p_i = \frac{1}{|\cH_{A'_i}|+1} \left(1 + \sum\limits_{h \in \cH_{A'_i}} \ind{h \geq H'_i}\right).
\end{equation}
Under the null hypothesis that $\cH_{A'_i}$ is exchangeable with $H'_i$, $p_i$ is a valid p-value (see Appendix~\ref{app:p-val} for a proof).
Furthermore, the dependence of these p-values asymptotically is negligible.

With this p-value in hand, we can test against this null hypothesis sequentially by using Fisher's combination test~\cite{fisher1928statistical} along with a variant of the Bonferroni correction.
In particular, for each user, after their $j$th vote, we compute $M_j = -2\sum\limits_{i=1}^j\log(p_i)$.
At 5 randomly chosen values of $j$ between 1 and 100, we identify a user as anomalous if $M_j \geq \chi^2_{2j, 1-\alpha/5}$. 
(The times are randomly chosen, as to avoid anomalous users strategizing to hack this p-value.)
Despite the heuristic application of this procedure, it seems to work well in our small-scale tests reported in Table~\ref{tab:confusion}.

\section{Data Analysis}



To examine whether Arena's crowdsourced data reflects real-world use cases, we conduct topic modeling on the user prompts.
We show how effective are these prompts in distinguishing models. Lastly, we validate the vote quality by relabeling data with experts.

\subsection{Topic Modeling on User Prompts}

To study the prompt diversity, we build a topic modeling pipeline with BERTopic\footnote{\url{https://github.com/MaartenGr/BERTopic}}~\cite{grootendorst2022bertopic}.
We start with transforming user prompts into representation vectors using OpenAI's text embedding model (text-embedding-3-small).
To mitigate the curse of dimensionality for data clustering, we employ UMAP (Uniform Manifold Approximation and Projection)~\citep{mcinnes2020umap} to reduce the embedding dimension from 1,536 to 5.
We then use the hierarchical density-based clustering algorithm, HDBSCAN, to identify topic clusters with minimum cluster size 32.
Finally, to obtain topic labels, we sample 10 prompts from each topic cluster and feed into GPT-4-Turbo for topic summarization.

\begin{figure}
\centering
\vspace{-5px}
\includegraphics[width=\linewidth]{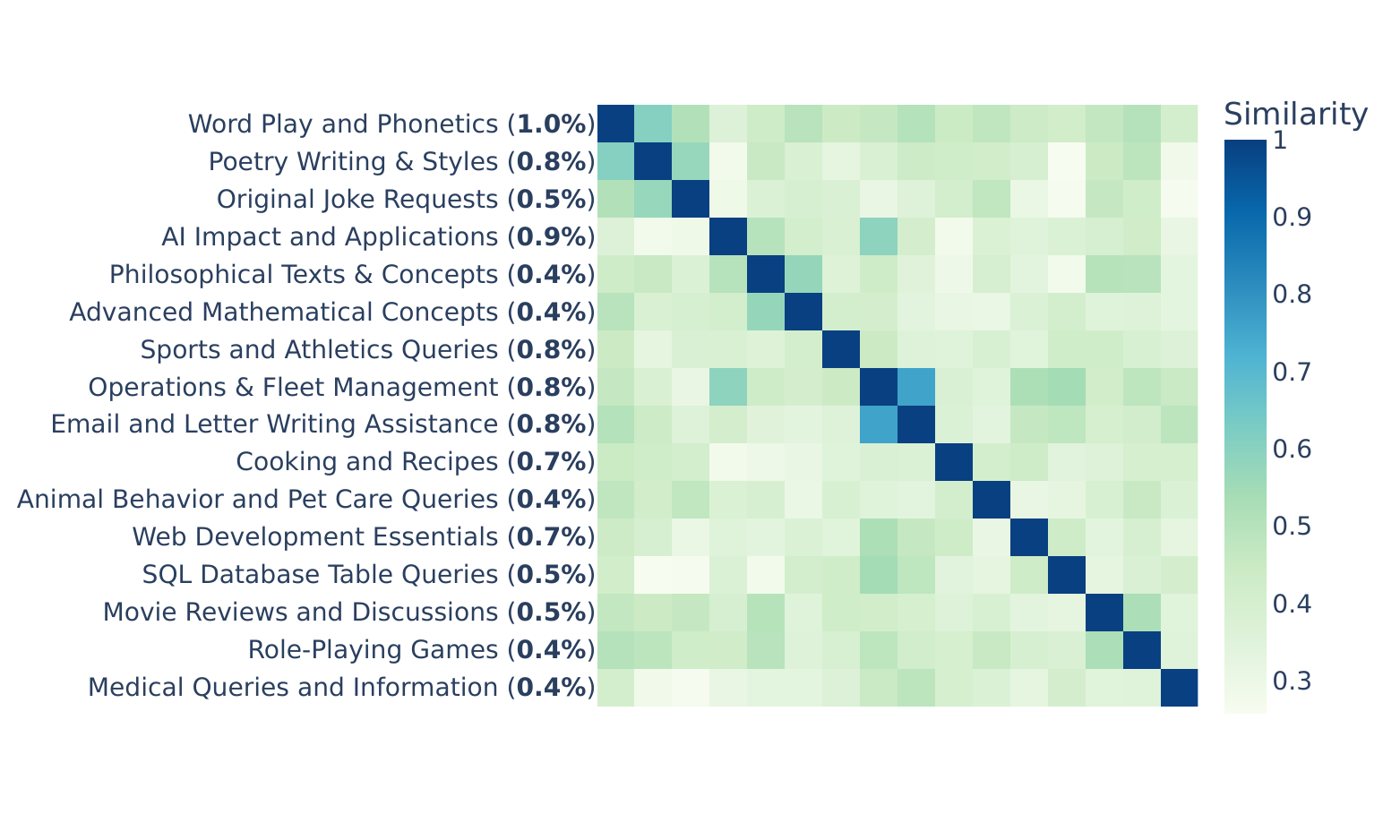}
\vspace{-40px}
\caption{Similarity matrix of top-16 topic clusters. The number followed by the topic label represents the cluster size in percentage. Note that similarity is computed by cluster's centroid embeddings, hence diagonals are always one.}
\vspace{-10px}
\label{fig:topic}
\end{figure}

The pipeline identifies 600 clusters covering a wide range of topics including poetry writing, coding, math, and medical queries.
We present the top-16 topic clusters in Figure~\ref{fig:topic}.
We observe that the largest cluster only accounts for 1\% of the entire set and the rest quickly drop to <0.5\%, and the similarity between clusters is small, showing a long-tail and diverse distribution.
Due to space limit, we present the similarity matrix and cluster hierarchy of top-64 clusters in Figure~\ref{fig:topic-64} and \ref{fig:hierarchy} in Appendix.

\subsection{Can Arena Prompts Distinguish Models?}

Next, we study how effective are these topic clusters in distinguishing models strengths.
Constructing challenging prompts has become increasingly difficult due to LLMs' fast growing capabilities. For example, open models such as Llama-2-70b-chat can likely answer inquiries about movie or travel recommendation as good as GPT-4, but not in other domains such as reasoning or coding. To demonstrate, we sample 30 prompts from seven topic clusters and compare the performance of Llama-2-70b-chat and GPT-4.
To control variables, we factor out user votes and consider LLM-as-judge~\cite{zheng2023judging} to evaluate model response.
Results are shown in Table~\ref{table:challenging-clusters}, where we see GPT-4 has significantly higher win-rate (up to 97\%) in clusters that require coding and reasoning skills. On the other hand, for clusters with less problem-solving tasks, GPT-4 win-rate drops to below 60\%. We show examples in Appendix~\ref{subsec:clusters-example}.
This result shows models may exhibit varying strengths in different areas, but also highlights some of the topic clusters in \system are effective in differentiate models.

\begin{table}[t!]
\centering
\vspace{-1em}
\caption{GPT-4-0613's win-rate against Llama-2-70b-chat on 30 sample prompts from various topic clusters. We use GPT-4-turbo as judge to evaluate model responses in pairwise comparison.}
\label{table:challenging-clusters}
\footnotesize
{\small
\begin{tabular}{lrr}
\toprule
Topic Cluster & Win-rate & Size\\
\midrule
Python Game Programming Challenge   & 96.7\% & 0.2\%\\
C/C++ Process Multi-Threading       & 86.7\% & 0.3\% \\
SQL Query Database Assistance       & 73.3\% & 0.2\%\\
Poetry Writing Prompts              & 66.7\% & 1.1\%\\
Python Coding Basics                & 65.0\% & 0.2\%\\
Linguistic Analysis \& Wordplay     & 58.3\% & 0.7\%\\
Travel Itinerary Planning           & 58.3\% & 0.4\% \\
Movie Recommendations \& Ratings    & 53.3\% & 0.2\% \\
\bottomrule
\end{tabular}}
\vspace{-5px}
\end{table}

\begin{figure}
    \centering
    \includegraphics[width=\linewidth]{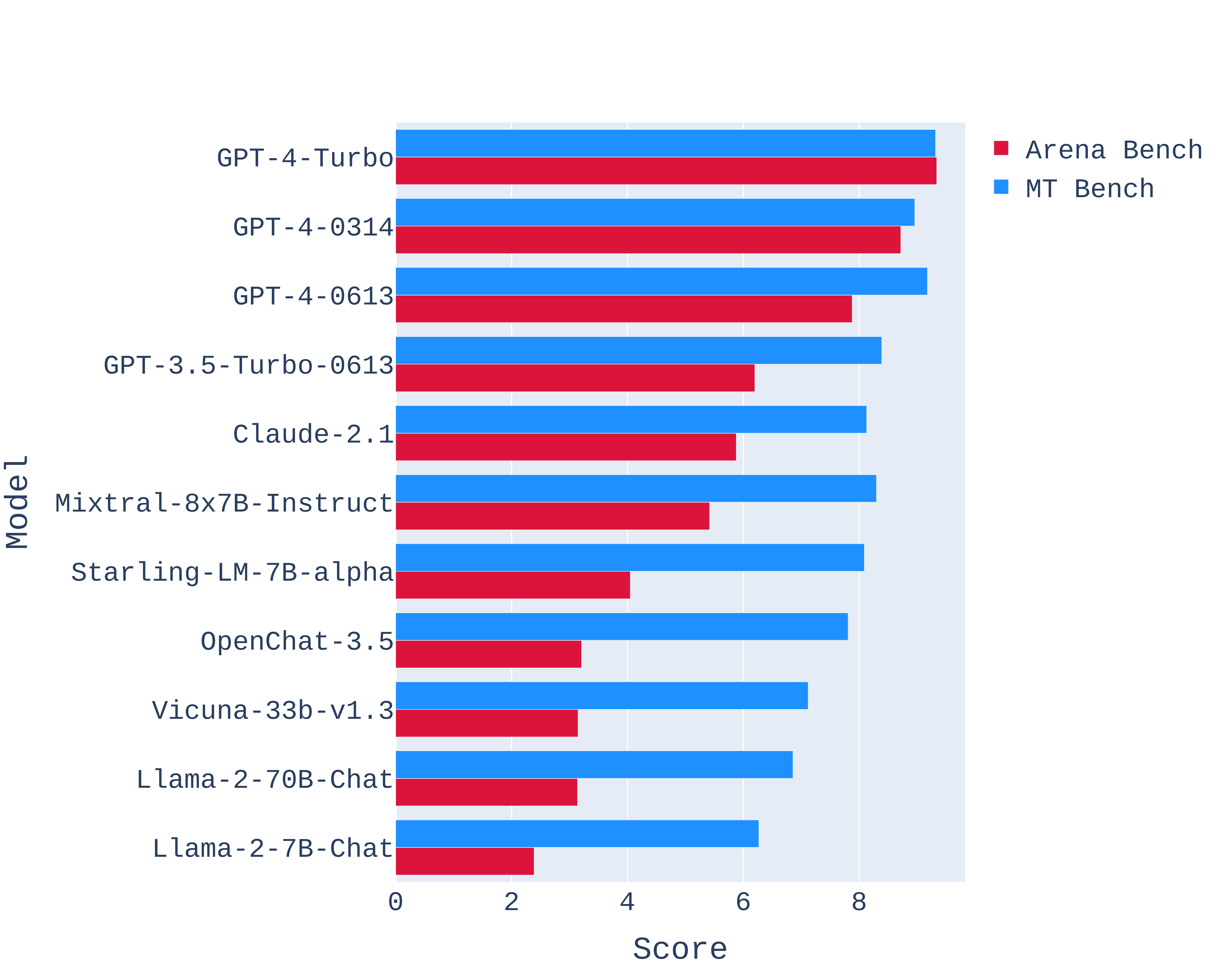}
    \vspace{-25px}
    \caption{Model's performance between Arena Bench and MT-Bench, showing an increased gap between open and proprietary models. Both uses GPT-4 as judge.}
    \vspace{-15px}
    \label{fig:arena_bench-vs-mt_bench}
\end{figure}

\textbf{Building Challenging Benchmark.} To further demonstrate the prompt quality, we show it is possible to construct a challenging benchmark with crowd-sourced user prompts.
To ensure both topic coverage and quality, we first run the topic modeling pipeline and follow a similar procedure in \citet{zheng2023lmsyschat1m} to select challenging questions sampled from each topic cluster.
Examples prompts and evaluation procedures can be found in the Appendix~\ref{subsec:arena-bench-examples} and Appendix~\ref{subsec:arena-bench-sys-prompt}, respectively.
We observe the selected prompts are highly effective in differentiating models. In Figure~\ref{fig:arena_bench-vs-mt_bench}, we compare Arena bench against a widely used LLM benchmark, MT-Bench~\cite{zheng2023judging}.
We can see that Arena Bench effectively reveals a significant gap in performance between proprietary and the strongest open models.

\subsection{Validating Vote Quality}

To assess the quality of crowdsourced votes, we randomly selected 160 battles between GPT-4-Turbo and Llama-2-13B, as well as GPT-4-Turbo and GPT-3.5-Turbo-0613.
We then asked experts\footnote{The laborers are graduate students at UC Berkeley.} to label their preference per comparison. The experts were given the prompts and answers blindly, and asked to carefully fact-check model's answer with external resources like search engine. Manually labeling each data point took on average 3-5 minutes. For reference, we also use GPT-4 as a judge for pairwise comparisons. The agreement rate between crowd-users, experts, and GPT-4-judge are presented in Table~\ref{table:agreement-rate}. The corresponsing win-rate are shown in Table~\ref{table:gpt-4-turbo-win-rate}.

To summarize, we observe high agreement rates (72\% to 83\%) between Arena crowd-user and experts in both setup. Note that agreement rates between two experts are around similar levels (79.4\% and 89.8\%). As for the 10\%-20\% disagreement between experts, it is mostly due to some user prompts don't have a ground truth answer. Depending on the preference of the evaluator, sometimes both answers can be argued as being better than the other one, such as the examples in Appendix ~\ref{subsec:no-ground-truth-example}.
The gap between crowd-vs-expert agreement rate and expert-vs-expert agreement rate (5\%-10\%) is mostly attributed to crowd user making mistakes or overlooking factual errors in model's response.
Overall, the agreement rates presented in Table~\ref{table:agreement-rate} validate the decent quality of crowd-sourced votes in \system.

\begin{table}[t!]
\centering
\caption{Pairwise agreement rate between crowd-user, gpt-4 judge, and experts on pairwise battles. The top part of the table is between GPT-4-Turbo and Llama-2-13b-chat. The bottom is between GPT-4-Turbo and GPT-3.5-Turbo-0613.}
\label{table:agreement-rate}
\footnotesize
\begin{tabular}{lrrr}
\toprule
Llama-2-13b & Expert 1 & Expert 2  & GPT-4 \\
\midrule
Crowd      & 72.8\% & 77.8\% & 75.6\% \\
Expert 1        &   -   & 89.8\% & 81.0\% \\
Expert 2        &   -   &   -       & 78.5\% \\
\midrule
GPT-3.5-Turbo & Expert 1 & Expert 2  & GPT-4 \\
\midrule
Crowd      & 73.8\%& 83.1\% & 75.6\%\\
Expert 1        &   -   & 79.4\% & 76.3\%\\
Expert 2        &   -   &   -    & 79.3\%\\
\bottomrule
\end{tabular}
\vspace{-15px}
\end{table}

\begin{table}[t!]
\centering
\caption{GPT-4-Turbo's win-rate across crowd-user, gpt-4 judge, and experts on pairwise battles against Llama-2-13b and GPT-3.5-Turbo-0613.}
\label{table:gpt-4-turbo-win-rate}
\resizebox{1\columnwidth}{!}{
\begin{tabular}{lrrrr}
\toprule
Baseline          & Arena User & Expert 1 & Expert 2 & GPT-4 \\
\midrule
Llama-2-13b     & 81.2\% & 89.4\% & 86.9\% & 78.8\% \\
GPT-3.5-Turbo   & 76.3\% & 82.5\% & 89.4\% & 79.4\% \\
\bottomrule
\end{tabular}}
\vspace{-0.5em}
\end{table}





\section{Experiments}

\subsection{Ranking system}

\begin{figure}
\centering
\vspace{-5px}
\includegraphics[width=0.48\textwidth]{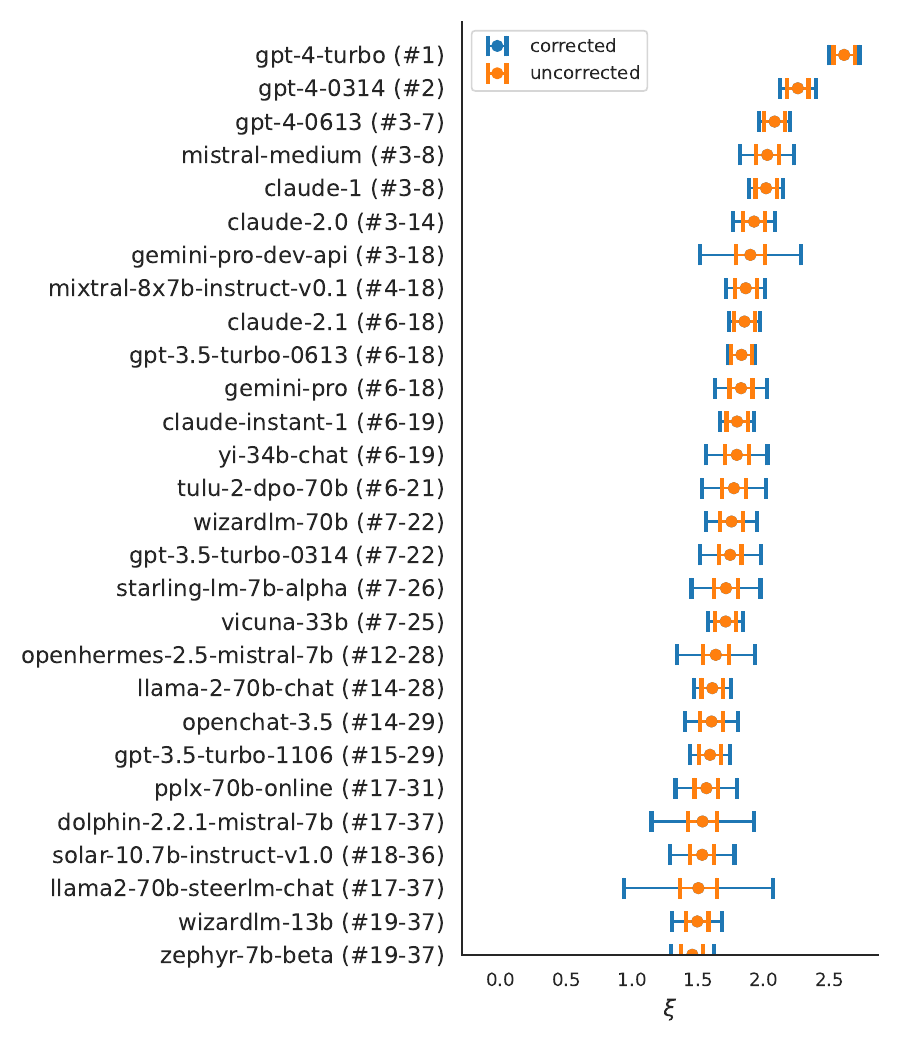}
\vspace{-30px}
 \caption{Intervals for the BT coefficients with and without multiplicity correction. The multiplicity correction, in this case a chi-square CLT interval, is technically required for the purpose of calculating the ranking, because it ensures \emph{all} scores are simultaneously contained in their intervals (and the ranking is a function of all the scores). However, it induces extra conservatism, so we report both intervals.}
\vspace{-5px}
\label{fig:sandwich-intervals}
\end{figure}

\textbf{Computing the rank on real data.} In this section, we report results from our experiments on approximate ranking.
For this experiment, we ran a replay of $T=213,576$ historical votes from our online platform and calculate the BT coefficients using our earlier-described estimation algorithm with confidence intervals; see Figure~\ref{fig:sandwich-intervals} for these intervals (with and without multiplicity correction; the formal notion of approximate ranking technically requires multiplicity correction, but it makes the intervals looser).

\textbf{Evaluating the coverage of the intervals.}
A natural follow-up question is whether or not the intervals are doing their job correctly: whether they cover the true BT coefficients with probability at least (and almost exactly) $1-\alpha$.
Of course, this cannot be evaluated on real data, so we run a simulation.
A vector of BT coefficients is drawn, with each coordinate sampled i.i.d.\ from a distribution $\mathsf{beta}(1/\gamma, 1/\gamma)$; we take $\gamma=2$ in Figure~\ref{fig:cvg-sim} (and we vary $\gamma$ in Appendix~\ref{app:simulation-study}).
Given these coefficients, a dataset is synthesized, and the coverage and average width are computed for each of 20 trials.
The results can be seen in Figure~\ref{fig:cvg-sim} for the uncorrected intervals
The coverage of the intervals behaves as expected, centering around $1-\alpha$, regardless of the number of models. 
Meanwhile, the more models are included, the larger the intervals become.
\begin{figure}
\centering
\includegraphics[width=0.48\textwidth]{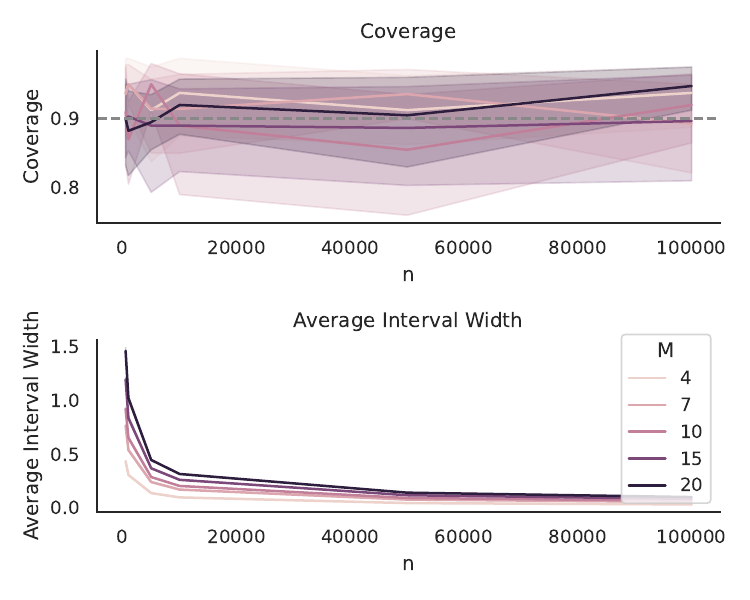}
 \vspace{-30px}
 \caption{Intervals for the BT coefficients as a function of the number of samples and the number of models $M$.}
\label{fig:cvg-sim}
\end{figure}

\begin{figure}
\centering
\includegraphics[width=0.9\linewidth]{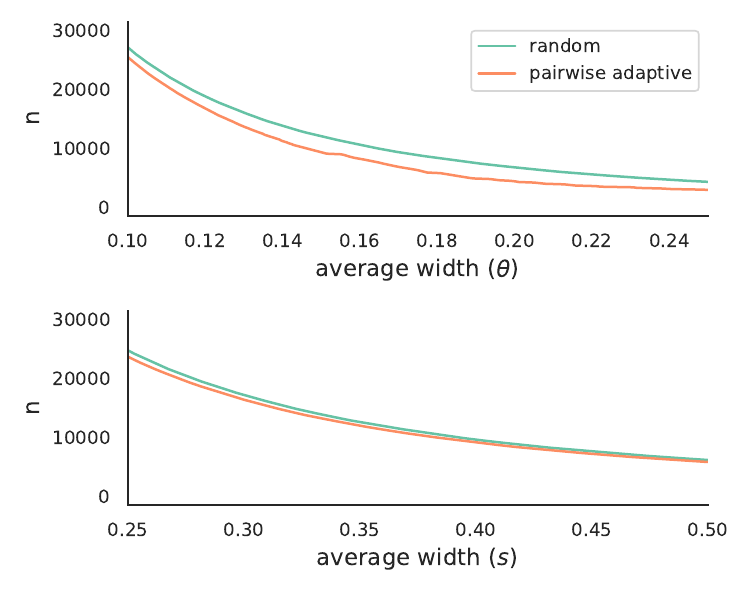}
 \vspace{-15px}
 \caption{Interval widths on the win matrix (upper figure) and on the BT coefficients (lower figure) as a function of the number of samples, for random sampling and also adaptive sampling. Improvements from adaptive sampling can be seen in both cases, although they are more subtle on the scale of the score. }
\label{fig:width_vs_samples_real}
\vspace{-15px}
\end{figure}

\textbf{Evaluating the active sampling rule.}
Next, we discuss the evaluation of our active sampling rule as Equation~\eqref{eq:adaptive} for win matrix estimation.
We evaluate this sampling rule by taking the best fit BT coefficients to our 213,576 point sized holdout set, and then sampling from that distribution using our active sampling algorithm.
The results are displayed in Figure~\ref{fig:width_vs_samples_real}.
It is hard to tell by looking at plots, but the improvement is substantial:
To estimate $\theta^*$ to a precision of 0.2, random needs 6,800 samples and adaptive needs 4,400 samples; meanwhile to estimate the score to a precision of 0.3, random needs 17,200 samples and adaptive needs 16,400 samples.
Thus, the random baseline requires 54\% and 5\% more data to achieve the same level of precision, respectively.
One can see from the plots in Figure~\ref{fig:width_vs_samples_real} that these results are not cherry-picked: the sample-efficiency of our method is better at all values on the horizontal axis.

\subsection{Anomalous Users Detection}

We evaluate the outlier detection method in Section~\ref{sec:outlier}.
We construct the evaluation set by manually identifying 25 anomalous users whose inputs are highly repetitive or meaningless (e.g., asking ``hi'' for 100 times or inputting garbled texts).
We randomly sample 25 normal users with at least 50 votes, and inspect their input prompts to ensure no abnormal behaviors.
As mentioned in Section~\ref{sec:outlier}, per user we compute five $M_j$ and identify the user as anomalous if $M_j \geq \chi^2_{2j, 1-\alpha/5}$.
We present results of two different $\alpha$ (i.e., the significance leval) in Table~\ref{tab:confusion}. We find the detection method effective (e.g., reaching 90\% true positive and 60-70\% true negative rate). We inspect the false negative errors and find those are from users do not always behave abnormally, making them harder to detect.

\begin{table}[]
\centering
\caption{Confusion matrix of different $\alpha$. ``Pred.'' means predicted. Positive means anomalous and negative means normal.} 
\label{tab:confusion}
\begin{tabular}{lrr}
\toprule
$\alpha=0.1$  & Pred. Positive & Pred. Negative \\
\midrule
Actual Positive  &  13/14  &   12/36      \\ 
Actual Negative  &  1/14  &   24/36    \\ 
\bottomrule
$\alpha=0.3$  & Pred. Positive & Pred. Negative \\
\midrule
Actual Positive  &  21/29  &   4/21      \\ 
Actual Negative  &  8/29  &    17/21     \\ 
\bottomrule
\end{tabular}
\vspace{-10px}
\end{table}

\section{Discussion}

\textbf{Limitations.} 
Although our user base is extensive, we anticipate that it will primarily consist of LLM hobbyists and researchers who are eager to experiment with and evaluate the latest LLMs. This inclination may result in a biased distribution of users. Additionally, despite the wide array of topics encompassed by the prompts discussed in previous sections, the data predominantly comes from our online chat interface. This source might not accurately reflect the real-world usage of LLMs in production environments or specialized domains, potentially leading to a skewed prompt distribution. Moreover, our study concentrates on assessing the helpfulness of LLMs but overlooks their safety aspects. We recognize the possibility and necessity of a parallel mechanism to evaluate the safety of these models.

\textbf{Future Directions.}
In our future work, we plan to develop comprehensive topic leaderboards and establish a dedicated section for multimodal and agent-based LLMs in more dynamic, gamified settings, catering to more complex tasks.
We also believe our approach to detecting harmful users could be improved and made more formally rigorous by using the theory of nonnegative supermartingales and E-values~\cite{howard2020time, waudby2020estimating, vovk2021values, ramdas2023game}; this would deal with the dependence, but the variants we tried did not perform well in terms of power.

\section{Conclusion}

In this paper, we present \system, an open platform for evaluating LLMs through crowdsourced, pairwise human preferences.
We conduct an in-depth analysis of the crowdsourced user prompts and preference votes to validate the diversity and quality.
We develop an efficient model sampling and ranking algorithm.
Our dataset including 100K pairwise preference votes will be released for future research.



\section*{Acknowledgments}
This project is supported by sponsorship from Kaggle, MBZUAI, a16z, Together AI, Anyscale, and HuggingFace.
This project is also partly supported by Accenture, AMD, Google, IBM, Intel, Microsoft, Samsung SDS, SAP, Uber, and VMware.
The authors would like to thank Siyuan Zhuang for insightful discussion and Tijana Zrni\'c for helpful feedback on the manuscript.

\nocite{langley00}

\bibliography{example_paper}

\begin{thebibliography}{52}
\providecommand{\natexlab}[1]{#1}
\providecommand{\url}[1]{\texttt{#1}}
\expandafter\ifx\csname urlstyle\endcsname\relax
  \providecommand{\doi}[1]{doi: #1}\else
  \providecommand{\doi}{doi: \begingroup \urlstyle{rm}\Url}\fi

\bibitem[Bai et~al.(2022)Bai, Jones, Ndousse, Askell, Chen, DasSarma, Drain, Fort, Ganguli, Henighan, et~al.]{bai2022training}
Bai, Y., Jones, A., Ndousse, K., Askell, A., Chen, A., DasSarma, N., Drain, D., Fort, S., Ganguli, D., Henighan, T., et~al.
\newblock Training a helpful and harmless assistant with reinforcement learning from human feedback.
\newblock \emph{arXiv preprint arXiv:2204.05862}, 2022.

\bibitem[Boubdir et~al.(2023)Boubdir, Kim, Ermis, Hooker, and Fadaee]{boubdir2023elo}
Boubdir, M., Kim, E., Ermis, B., Hooker, S., and Fadaee, M.
\newblock Elo uncovered: Robustness and best practices in language model evaluation, 2023.

\bibitem[Bradley \& Terry(1952)Bradley and Terry]{bradley1952rank}
Bradley, R.~A. and Terry, M.~E.
\newblock Rank analysis of incomplete block designs: I. the method of paired comparisons.
\newblock \emph{Biometrika}, 39\penalty0 (3/4):\penalty0 324--345, 1952.

\bibitem[Busa-Fekete et~al.(2014{\natexlab{a}})Busa-Fekete, Huellermeier, and Szörényi]{pmlr-v32-busa-fekete14}
Busa-Fekete, R., Huellermeier, E., and Szörényi, B.
\newblock Preference-based rank elicitation using statistical models: The case of mallows.
\newblock In Xing, E.~P. and Jebara, T. (eds.), \emph{Proceedings of the 31st International Conference on Machine Learning}, volume~32 of \emph{Proceedings of Machine Learning Research}, pp.\  1071--1079, Bejing, China, 22--24 Jun 2014{\natexlab{a}}. PMLR.
\newblock URL \url{https://proceedings.mlr.press/v32/busa-fekete14.html}.

\bibitem[Busa-Fekete et~al.(2014{\natexlab{b}})Busa-Fekete, Huellermeier, and Szörényi]{preference_rank_elicitation}
Busa-Fekete, R., Huellermeier, E., and Szörényi, B.
\newblock Preference-based rank elicitation using statistical models: The case of mallows.
\newblock In Xing, E.~P. and Jebara, T. (eds.), \emph{Proceedings of the 31st International Conference on Machine Learning}, volume~32 of \emph{Proceedings of Machine Learning Research}, pp.\  1071--1079, Bejing, China, 22--24 Jun 2014{\natexlab{b}}. PMLR.
\newblock URL \url{https://proceedings.mlr.press/v32/busa-fekete14.html}.

\bibitem[Chen et~al.(2021)Chen, Tworek, Jun, Yuan, Pinto, Kaplan, Edwards, Burda, Joseph, Brockman, et~al.]{chen2021evaluating}
Chen, M., Tworek, J., Jun, H., Yuan, Q., Pinto, H. P. d.~O., Kaplan, J., Edwards, H., Burda, Y., Joseph, N., Brockman, G., et~al.
\newblock Evaluating large language models trained on code.
\newblock \emph{arXiv preprint arXiv:2107.03374}, 2021.

\bibitem[Chernoff(1992)]{sequential_design}
Chernoff, H.
\newblock \emph{Sequential Design of Experiments}, pp.\  345--360.
\newblock Springer New York, New York, NY, 1992.
\newblock ISBN 978-1-4612-4380-9.
\newblock \doi{10.1007/978-1-4612-4380-9_27}.
\newblock URL \url{https://doi.org/10.1007/978-1-4612-4380-9_27}.

\bibitem[Chiang \& Lee(2023)Chiang and Lee]{chiang-lee-2023-large}
Chiang, C.-H. and Lee, H.-y.
\newblock Can large language models be an alternative to human evaluations?
\newblock In Rogers, A., Boyd-Graber, J., and Okazaki, N. (eds.), \emph{Proceedings of the 61st Annual Meeting of the Association for Computational Linguistics (Volume 1: Long Papers)}, pp.\  15607--15631, Toronto, Canada, July 2023. Association for Computational Linguistics.
\newblock \doi{10.18653/v1/2023.acl-long.870}.
\newblock URL \url{https://aclanthology.org/2023.acl-long.870}.

\bibitem[Cobbe et~al.(2021)Cobbe, Kosaraju, Bavarian, Chen, Jun, Kaiser, Plappert, Tworek, Hilton, Nakano, et~al.]{cobbe2021gsm}
Cobbe, K., Kosaraju, V., Bavarian, M., Chen, M., Jun, H., Kaiser, L., Plappert, M., Tworek, J., Hilton, J., Nakano, R., et~al.
\newblock Training verifiers to solve math word problems.
\newblock \emph{arXiv preprint arXiv:2110.14168}, 2021.

\bibitem[Cui et~al.(2023)Cui, Yuan, Ding, Yao, Zhu, Ni, Xie, Liu, and Sun]{cui2023ultrafeedback}
Cui, G., Yuan, L., Ding, N., Yao, G., Zhu, W., Ni, Y., Xie, G., Liu, Z., and Sun, M.
\newblock Ultrafeedback: Boosting language models with high-quality feedback, 2023.

\bibitem[DiCiccio \& Efron(1996)DiCiccio and Efron]{diciccio1996bootstrap}
DiCiccio, T.~J. and Efron, B.
\newblock Bootstrap confidence intervals.
\newblock \emph{Statistical science}, 11\penalty0 (3):\penalty0 189--228, 1996.

\bibitem[Durrett(2019)]{durrett2019probability}
Durrett, R.
\newblock \emph{Probability: theory and examples}, volume~49.
\newblock Cambridge university press, 2019.

\bibitem[Elo(1967)]{elo1967proposed}
Elo, A.~E.
\newblock The proposed uscf rating system, its development, theory, and applications.
\newblock \emph{Chess Life}, 22\penalty0 (8):\penalty0 242--247, 1967.

\bibitem[Fisher(1928)]{fisher1928statistical}
Fisher, R.~A.
\newblock \emph{Statistical methods for research workers}.
\newblock Number~5. Oliver and Boyd, 1928.

\bibitem[Freedman(2006)]{freedman2006so}
Freedman, D.~A.
\newblock On the so-called ``huber sandwich estimator''' and ``robust standard errors'''.
\newblock \emph{The American Statistician}, 60\penalty0 (4):\penalty0 299--302, 2006.

\bibitem[Gemini et~al.(2023)Gemini, Anil, Borgeaud, Wu, Alayrac, Yu, Soricut, Schalkwyk, Dai, Hauth, et~al.]{team2023gemini}
Gemini, T., Anil, R., Borgeaud, S., Wu, Y., Alayrac, J.-B., Yu, J., Soricut, R., Schalkwyk, J., Dai, A.~M., Hauth, A., et~al.
\newblock Gemini: a family of highly capable multimodal models.
\newblock \emph{arXiv preprint arXiv:2312.11805}, 2023.

\bibitem[Geng et~al.(2023)Geng, Gudibande, Liu, Wallace, Abbeel, Levine, and Song]{koala_blogpost_2023}
Geng, X., Gudibande, A., Liu, H., Wallace, E., Abbeel, P., Levine, S., and Song, D.
\newblock Koala: A dialogue model for academic research.
\newblock Blog post, April 2023.
\newblock URL \url{https://bair.berkeley.edu/blog/2023/04/03/koala/}.

\bibitem[Grootendorst(2022)]{grootendorst2022bertopic}
Grootendorst, M.
\newblock Bertopic: Neural topic modeling with a class-based tf-idf procedure.
\newblock \emph{arXiv preprint arXiv:2203.05794}, 2022.

\bibitem[Hendrycks et~al.(2020)Hendrycks, Burns, Basart, Zou, Mazeika, Song, and Steinhardt]{hendrycks2020measuring}
Hendrycks, D., Burns, C., Basart, S., Zou, A., Mazeika, M., Song, D., and Steinhardt, J.
\newblock Measuring massive multitask language understanding.
\newblock In \emph{International Conference on Learning Representations}, 2020.

\bibitem[Howard et~al.(2020)Howard, Ramdas, McAuliffe, and Sekhon]{howard2020time}
Howard, S.~R., Ramdas, A., McAuliffe, J., and Sekhon, J.
\newblock Time-uniform chernoff bounds via nonnegative supermartingales.
\newblock 2020.

\bibitem[Huang et~al.(2023)Huang, Lin, Liu, Gong, Lu, Lei, Liang, Shen, Lin, Duan, et~al.]{huang2023competition}
Huang, Y., Lin, Z., Liu, X., Gong, Y., Lu, S., Lei, F., Liang, Y., Shen, Y., Lin, C., Duan, N., et~al.
\newblock Competition-level problems are effective llm evaluators.
\newblock \emph{arXiv preprint arXiv:2312.02143}, 2023.

\bibitem[Huber et~al.(1967)]{huber1967behavior}
Huber, P.~J. et~al.
\newblock The behavior of maximum likelihood estimates under nonstandard conditions.
\newblock In \emph{Proceedings of the fifth Berkeley symposium on mathematical statistics and probability}, volume~1, pp.\  221--233. Berkeley, CA: University of California Press, 1967.

\bibitem[Hunter(2004)]{mm_bradley_terry}
Hunter, D.~R.
\newblock {MM algorithms for generalized Bradley-Terry models}.
\newblock \emph{The Annals of Statistics}, 32\penalty0 (1):\penalty0 384 -- 406, 2004.
\newblock \doi{10.1214/aos/1079120141}.
\newblock URL \url{https://doi.org/10.1214/aos/1079120141}.

\bibitem[Karimi et~al.(2021)Karimi, G{\"u}rel, Karla{\v{s}}, Rausch, Zhang, and Krause]{karimi2021online}
Karimi, M.~R., G{\"u}rel, N.~M., Karla{\v{s}}, B., Rausch, J., Zhang, C., and Krause, A.
\newblock Online active model selection for pre-trained classifiers.
\newblock In \emph{International Conference on Artificial Intelligence and Statistics}, pp.\  307--315. PMLR, 2021.

\bibitem[Karpinska et~al.(2021)Karpinska, Akoury, and Iyyer]{karpinska-etal-2021-perils}
Karpinska, M., Akoury, N., and Iyyer, M.
\newblock The perils of using {M}echanical {T}urk to evaluate open-ended text generation.
\newblock In Moens, M.-F., Huang, X., Specia, L., and Yih, S. W.-t. (eds.), \emph{Proceedings of the 2021 Conference on Empirical Methods in Natural Language Processing}, pp.\  1265--1285, Online and Punta Cana, Dominican Republic, November 2021. Association for Computational Linguistics.
\newblock \doi{10.18653/v1/2021.emnlp-main.97}.
\newblock URL \url{https://aclanthology.org/2021.emnlp-main.97}.

\bibitem[Kiela et~al.(2021)Kiela, Bartolo, Nie, Kaushik, Geiger, Wu, Vidgen, Prasad, Singh, Ringshia, et~al.]{kiela2021dynabench}
Kiela, D., Bartolo, M., Nie, Y., Kaushik, D., Geiger, A., Wu, Z., Vidgen, B., Prasad, G., Singh, A., Ringshia, P., et~al.
\newblock Dynabench: Rethinking benchmarking in nlp.
\newblock In \emph{Proceedings of the 2021 Conference of the North American Chapter of the Association for Computational Linguistics: Human Language Technologies}, pp.\  4110--4124, 2021.

\bibitem[K{\"o}pf et~al.(2023)K{\"o}pf, Kilcher, von R{\"u}tte, Anagnostidis, Tam, Stevens, Barhoum, Duc, Stanley, Nagyfi, et~al.]{kopf2023openassistant}
K{\"o}pf, A., Kilcher, Y., von R{\"u}tte, D., Anagnostidis, S., Tam, Z.-R., Stevens, K., Barhoum, A., Duc, N.~M., Stanley, O., Nagyfi, R., et~al.
\newblock Openassistant conversations--democratizing large language model alignment.
\newblock \emph{arXiv preprint arXiv:2304.07327}, 2023.

\bibitem[Langley(2000)]{langley00}
Langley, P.
\newblock Crafting papers on machine learning.
\newblock In Langley, P. (ed.), \emph{Proceedings of the 17th International Conference on Machine Learning (ICML 2000)}, pp.\  1207--1216, Stanford, CA, 2000. Morgan Kaufmann.

\bibitem[Li et~al.(2023)Li, Zhang, Dubois, Taori, Gulrajani, Guestrin, Liang, and Hashimoto]{alpaca_eval}
Li, X., Zhang, T., Dubois, Y., Taori, R., Gulrajani, I., Guestrin, C., Liang, P., and Hashimoto, T.~B.
\newblock Alpacaeval: An automatic evaluator of instruction-following models.
\newblock \url{https://github.com/tatsu-lab/alpaca_eval}, 2023.

\bibitem[Li et~al.(2022)Li, Choi, Chung, Kushman, Schrittwieser, Leblond, Eccles, Keeling, Gimeno, Dal~Lago, et~al.]{li2022competition}
Li, Y., Choi, D., Chung, J., Kushman, N., Schrittwieser, J., Leblond, R., Eccles, T., Keeling, J., Gimeno, F., Dal~Lago, A., et~al.
\newblock Competition-level code generation with alphacode.
\newblock \emph{Science}, 378\penalty0 (6624):\penalty0 1092--1097, 2022.

\bibitem[Liang et~al.(2022)Liang, Bommasani, Lee, Tsipras, Soylu, Yasunaga, Zhang, Narayanan, Wu, Kumar, et~al.]{liang2022holistic}
Liang, P., Bommasani, R., Lee, T., Tsipras, D., Soylu, D., Yasunaga, M., Zhang, Y., Narayanan, D., Wu, Y., Kumar, A., et~al.
\newblock Holistic evaluation of language models.
\newblock \emph{arXiv preprint arXiv:2211.09110}, 2022.

\bibitem[Lin et~al.(2023)Lin, Wang, Tong, Wang, Guo, Wang, and Shang]{toxicchat}
Lin, Z., Wang, Z., Tong, Y., Wang, Y., Guo, Y., Wang, Y., and Shang, J.
\newblock {T}oxic{C}hat: Unveiling hidden challenges of toxicity detection in real-world user-{AI} conversation.
\newblock In Bouamor, H., Pino, J., and Bali, K. (eds.), \emph{Findings of the Association for Computational Linguistics: EMNLP 2023}, pp.\  4694--4702, Singapore, December 2023. Association for Computational Linguistics.
\newblock \doi{10.18653/v1/2023.findings-emnlp.311}.
\newblock URL \url{https://aclanthology.org/2023.findings-emnlp.311}.

\bibitem[Liu et~al.(2009)]{liu2009learning}
Liu, T.-Y. et~al.
\newblock Learning to rank for information retrieval.
\newblock \emph{Foundations and Trends{\textregistered} in Information Retrieval}, 3\penalty0 (3):\penalty0 225--331, 2009.

\bibitem[McInnes et~al.(2020)McInnes, Healy, and Melville]{mcinnes2020umap}
McInnes, L., Healy, J., and Melville, J.
\newblock Umap: Uniform manifold approximation and projection for dimension reduction, 2020.

\bibitem[OpenAI(2023)]{openai2023gpt4}
OpenAI.
\newblock Gpt-4 technical report.
\newblock \emph{arXiv preprint arXiv:2303.08774}, 2023.

\bibitem[Oren et~al.(2023)Oren, Meister, Chatterji, Ladhak, and Hashimoto]{oren2023proving}
Oren, Y., Meister, N., Chatterji, N., Ladhak, F., and Hashimoto, T.~B.
\newblock Proving test set contamination in black box language models.
\newblock \emph{arXiv preprint arXiv:2310.17623}, 2023.

\bibitem[Ouyang et~al.(2022)Ouyang, Wu, Jiang, Almeida, Wainwright, Mishkin, Zhang, Agarwal, Slama, Ray, Schulman, Hilton, Kelton, Miller, Simens, Askell, Welinder, Christiano, Leike, and Lowe]{ouyang2022training}
Ouyang, L., Wu, J., Jiang, X., Almeida, D., Wainwright, C.~L., Mishkin, P., Zhang, C., Agarwal, S., Slama, K., Ray, A., Schulman, J., Hilton, J., Kelton, F., Miller, L., Simens, M., Askell, A., Welinder, P., Christiano, P., Leike, J., and Lowe, R.
\newblock Training language models to follow instructions with human feedback, 2022.

\bibitem[Ramdas et~al.(2023)Ramdas, Gr{\"u}nwald, Vovk, and Shafer]{ramdas2023game}
Ramdas, A., Gr{\"u}nwald, P., Vovk, V., and Shafer, G.
\newblock Game-theoretic statistics and safe anytime-valid inference.
\newblock \emph{Statistical Science}, 38\penalty0 (4):\penalty0 576--601, 2023.

\bibitem[Rao \& Kupper(1967)Rao and Kupper]{tie_in_bradley_terry}
Rao, P.~V. and Kupper, L.~L.
\newblock Ties in paired-comparison experiments: A generalization of the bradley-terry model.
\newblock \emph{Journal of the American Statistical Association}, 62\penalty0 (317):\penalty0 194--204, 1967.
\newblock \doi{10.1080/01621459.1967.10482901}.

\bibitem[Srivastava et~al.(2023)Srivastava, Rastogi, Rao, Shoeb, Abid, Fisch, Brown, Santoro, Gupta, Garriga-Alonso, et~al.]{srivastava2023beyond}
Srivastava, A., Rastogi, A., Rao, A., Shoeb, A. A.~M., Abid, A., Fisch, A., Brown, A.~R., Santoro, A., Gupta, A., Garriga-Alonso, A., et~al.
\newblock Beyond the imitation game: Quantifying and extrapolating the capabilities of language models.
\newblock \emph{Transactions on Machine Learning Research}, 2023.

\bibitem[Sz\"{o}r\'{e}nyi et~al.(2015)Sz\"{o}r\'{e}nyi, Busa-Fekete, Paul, and H\"{u}llermeier]{online_rank}
Sz\"{o}r\'{e}nyi, B., Busa-Fekete, R., Paul, A., and H\"{u}llermeier, E.
\newblock Online rank elicitation for plackett-luce: A dueling bandits approach.
\newblock In Cortes, C., Lawrence, N., Lee, D., Sugiyama, M., and Garnett, R. (eds.), \emph{Advances in Neural Information Processing Systems}, volume~28. Curran Associates, Inc., 2015.
\newblock URL \url{https://proceedings.neurips.cc/paper_files/paper/2015/file/7eacb532570ff6858afd2723755ff790-Paper.pdf}.

\bibitem[Touvron et~al.(2023)Touvron, Martin, Stone, Albert, Almahairi, Babaei, Bashlykov, Batra, Bhargava, Bhosale, et~al.]{touvron2023llama2}
Touvron, H., Martin, L., Stone, K., Albert, P., Almahairi, A., Babaei, Y., Bashlykov, N., Batra, S., Bhargava, P., Bhosale, S., et~al.
\newblock Llama 2: Open foundation and fine-tuned chat models.
\newblock \emph{arXiv preprint arXiv:2307.09288}, 2023.

\bibitem[Vovk \& Wang(2021)Vovk and Wang]{vovk2021values}
Vovk, V. and Wang, R.
\newblock E-values: Calibration, combination and applications.
\newblock \emph{The Annals of Statistics}, 49\penalty0 (3):\penalty0 1736--1754, 2021.

\bibitem[Wang et~al.(2023)Wang, Kordi, Mishra, Liu, Smith, Khashabi, and Hajishirzi]{wang-etal-2023-self-instruct}
Wang, Y., Kordi, Y., Mishra, S., Liu, A., Smith, N.~A., Khashabi, D., and Hajishirzi, H.
\newblock Self-instruct: Aligning language models with self-generated instructions.
\newblock In Rogers, A., Boyd-Graber, J., and Okazaki, N. (eds.), \emph{Proceedings of the 61st Annual Meeting of the Association for Computational Linguistics (Volume 1: Long Papers)}, pp.\  13484--13508, Toronto, Canada, July 2023. Association for Computational Linguistics.
\newblock \doi{10.18653/v1/2023.acl-long.754}.
\newblock URL \url{https://aclanthology.org/2023.acl-long.754}.

\bibitem[Waudby-Smith \& Ramdas(2020)Waudby-Smith and Ramdas]{waudby2020estimating}
Waudby-Smith, I. and Ramdas, A.
\newblock Estimating means of bounded random variables by betting.
\newblock \emph{arXiv preprint arXiv:2010.09686}, 2020.

\bibitem[White(1982)]{white1982maximum}
White, H.
\newblock Maximum likelihood estimation of misspecified models.
\newblock \emph{Econometrica: Journal of the econometric society}, pp.\  1--25, 1982.

\bibitem[Yang et~al.(2023)Yang, Chiang, Zheng, Gonzalez, and Stoica]{yang2023rethinking}
Yang, S., Chiang, W.-L., Zheng, L., Gonzalez, J.~E., and Stoica, I.
\newblock Rethinking benchmark and contamination for language models with rephrased samples.
\newblock \emph{arXiv preprint arXiv:2311.04850}, 2023.

\bibitem[Zellers et~al.(2019)Zellers, Holtzman, Bisk, Farhadi, and Choi]{zellers2019hellaswag}
Zellers, R., Holtzman, A., Bisk, Y., Farhadi, A., and Choi, Y.
\newblock Hellaswag: Can a machine really finish your sentence?
\newblock In \emph{Proceedings of the 57th Annual Meeting of the Association for Computational Linguistics}, pp.\  4791--4800, 2019.

\bibitem[Zheng et~al.(2023{\natexlab{a}})Zheng, Chiang, Sheng, Li, Zhuang, Wu, Zhuang, Li, Lin, Xing, Gonzalez, Stoica, and Zhang]{zheng2023lmsyschat1m}
Zheng, L., Chiang, W.-L., Sheng, Y., Li, T., Zhuang, S., Wu, Z., Zhuang, Y., Li, Z., Lin, Z., Xing, E.~P., Gonzalez, J.~E., Stoica, I., and Zhang, H.
\newblock Lmsys-chat-1m: A large-scale real-world llm conversation dataset, 2023{\natexlab{a}}.

\bibitem[Zheng et~al.(2023{\natexlab{b}})Zheng, Chiang, Sheng, Zhuang, Wu, Zhuang, Lin, Li, Li, Xing, Zhang, Gonzalez, and Stoica]{zheng2023judging}
Zheng, L., Chiang, W.-L., Sheng, Y., Zhuang, S., Wu, Z., Zhuang, Y., Lin, Z., Li, Z., Li, D., Xing, E., Zhang, H., Gonzalez, J.~E., and Stoica, I.
\newblock Judging {LLM}-as-a-judge with {MT}-bench and chatbot arena.
\newblock In \emph{Thirty-seventh Conference on Neural Information Processing Systems Datasets and Benchmarks Track}, 2023{\natexlab{b}}.
\newblock URL \url{https://openreview.net/forum?id=uccHPGDlao}.

\bibitem[Zhong et~al.(2023)Zhong, Cui, Guo, Liang, Lu, Wang, Saied, Chen, and Duan]{zhong2023agieval}
Zhong, W., Cui, R., Guo, Y., Liang, Y., Lu, S., Wang, Y., Saied, A., Chen, W., and Duan, N.
\newblock Agieval: A human-centric benchmark for evaluating foundation models.
\newblock \emph{arXiv preprint arXiv:2304.06364}, 2023.

\bibitem[Zhu et~al.(2023)Zhu, Frick, Wu, Zhu, and Jiao]{starling2023}
Zhu, B., Frick, E., Wu, T., Zhu, H., and Jiao, J.
\newblock Starling-7b: Improving llm helpfulness \& harmlessness with rlaif, November 2023.

\end{thebibliography}
\bibliographystyle{icml2023}

\newpage
\appendix
\onecolumn


\begin{figure}
\centering
\includegraphics[width=\linewidth]{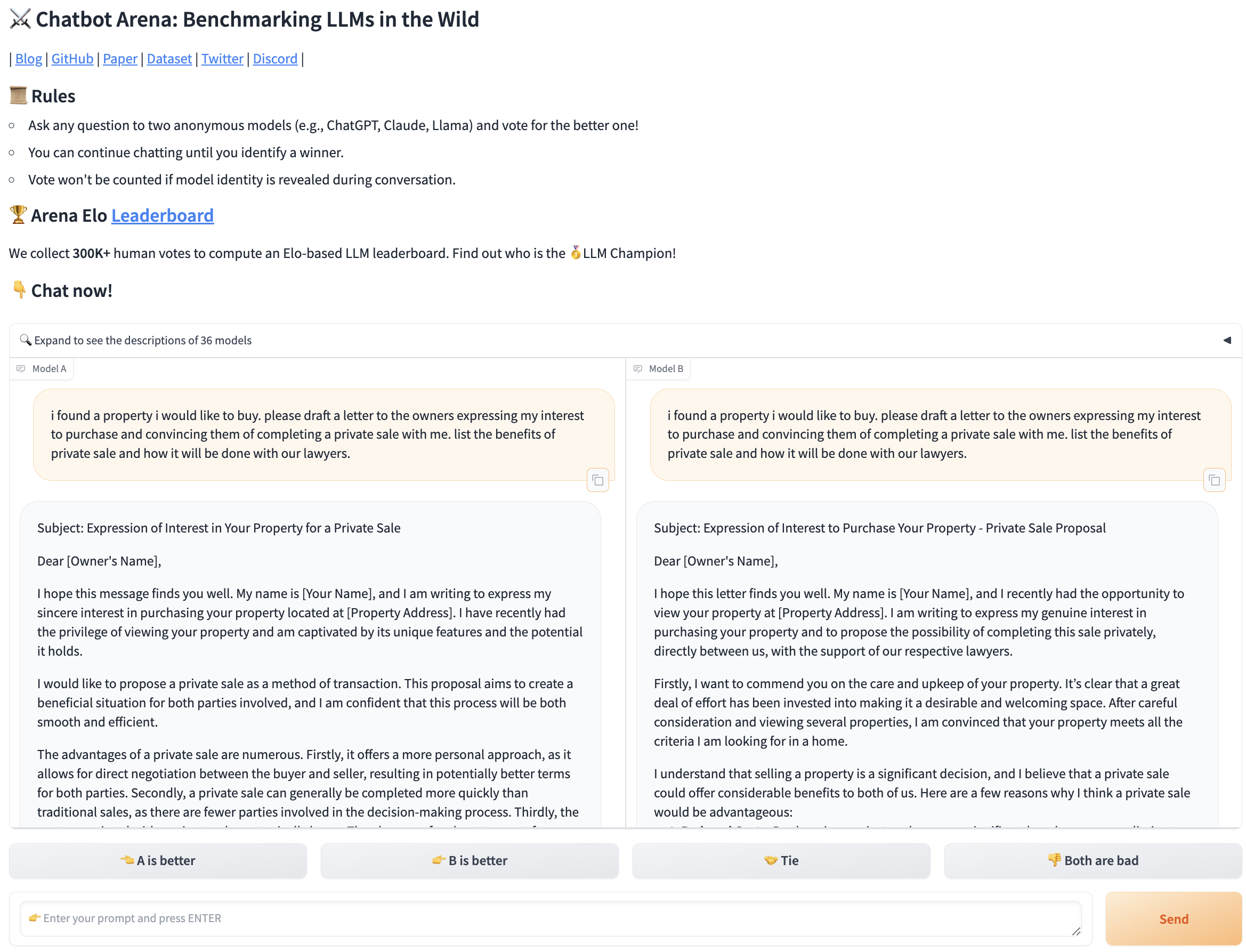}
\caption{Screenshot of \system.}
\label{fig:screenshot-arena}
\end{figure}

\begin{figure}
\centering
\vspace{-100px}
\includegraphics[width=0.95\textwidth]{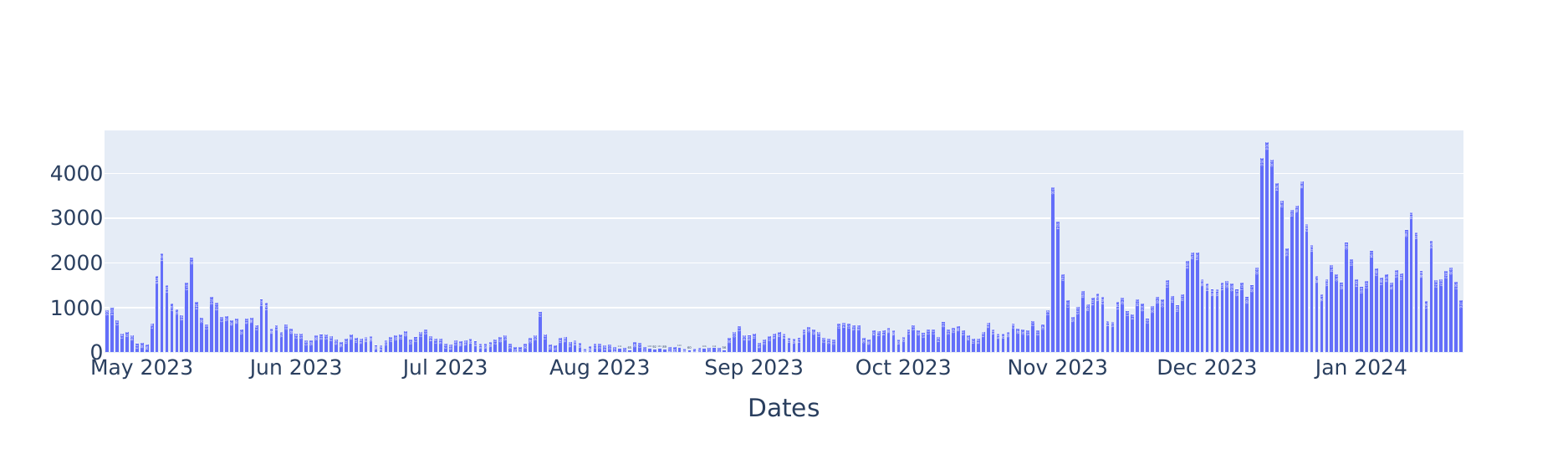}
\vspace{-20px}
\caption{The number of votes over time}
\label{fig:daily-vote}
\end{figure}

\begin{figure}
    \centering
    \includegraphics[width=0.95\linewidth]{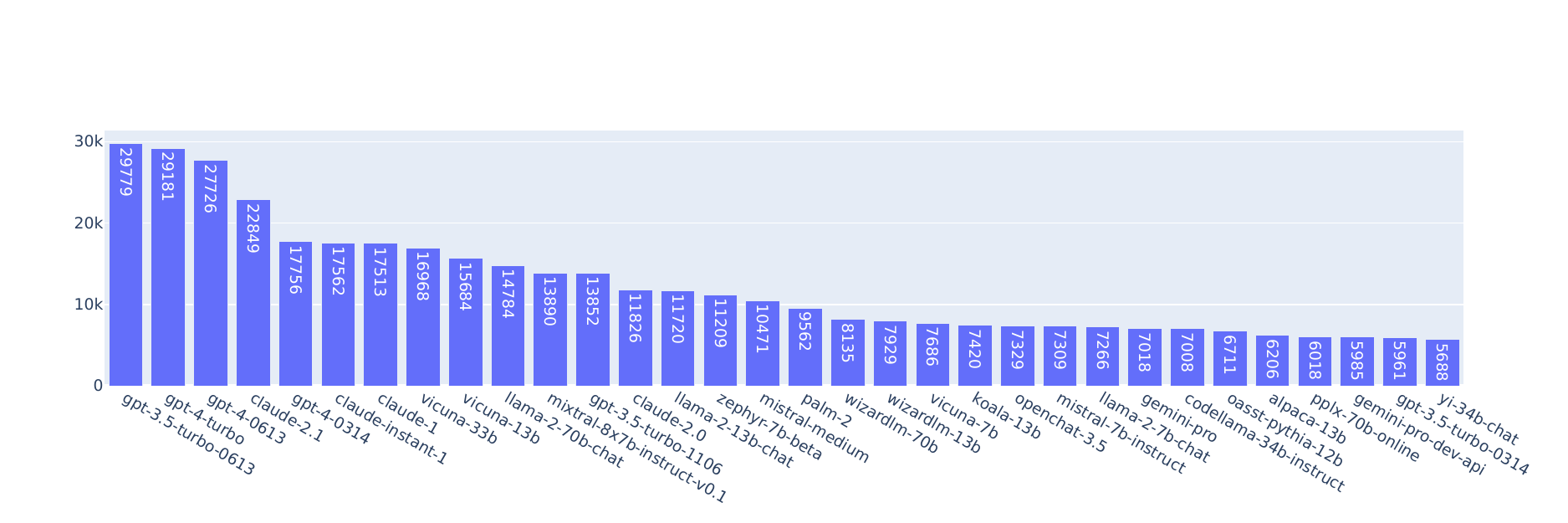}
    \vspace{-20px}
    \caption{The number of votes per model.}
    \label{fig:vote-cnt}
\end{figure}

\newpage

\begin{figure}[H]
\centering
\includegraphics[width=\linewidth]{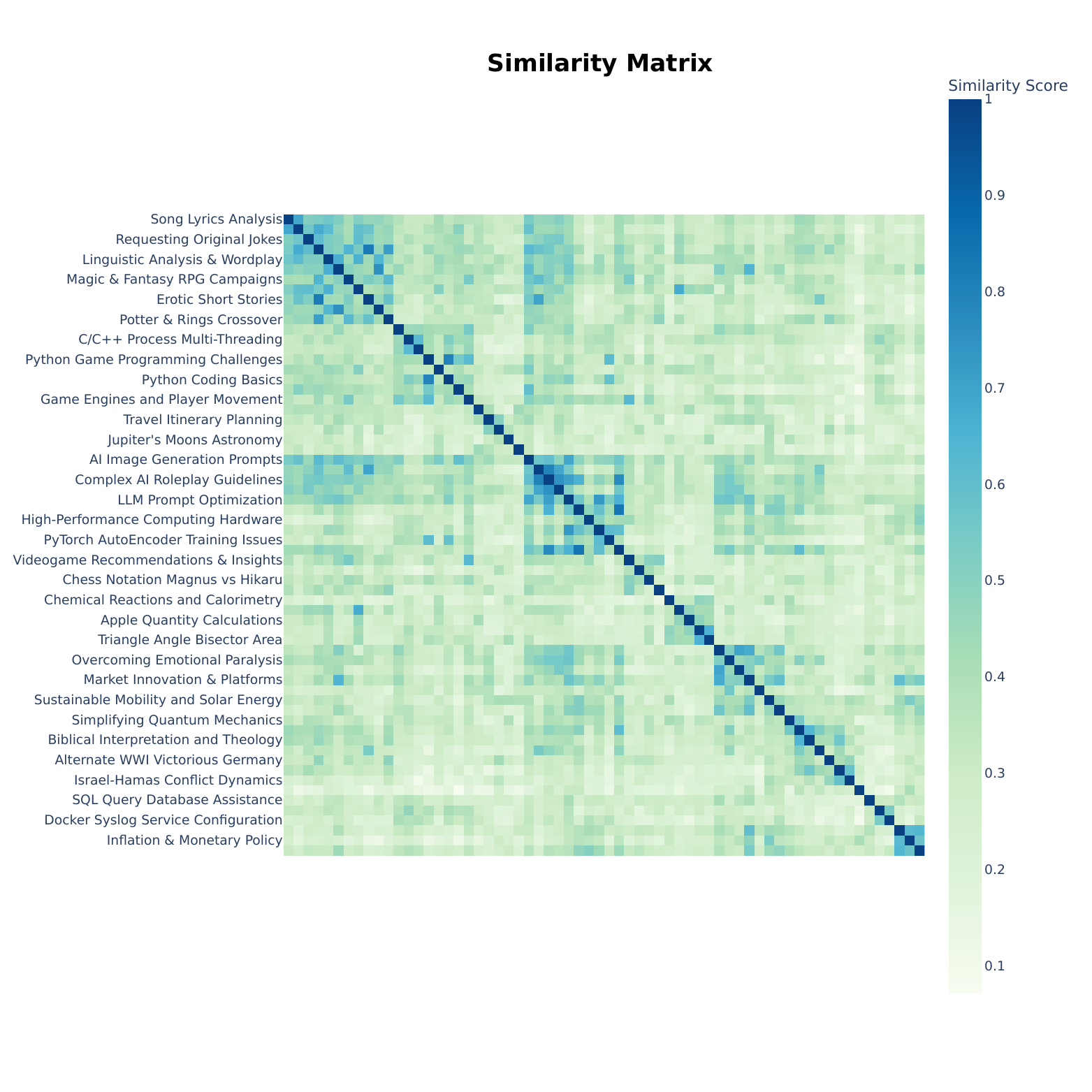}
\vspace{-50px}
\caption{Similarity matrix of top-64 topic clusters.}
\label{fig:topic-64}
\end{figure}

\newpage

\begin{figure}[H]
\centering
\includegraphics[width=\textwidth]{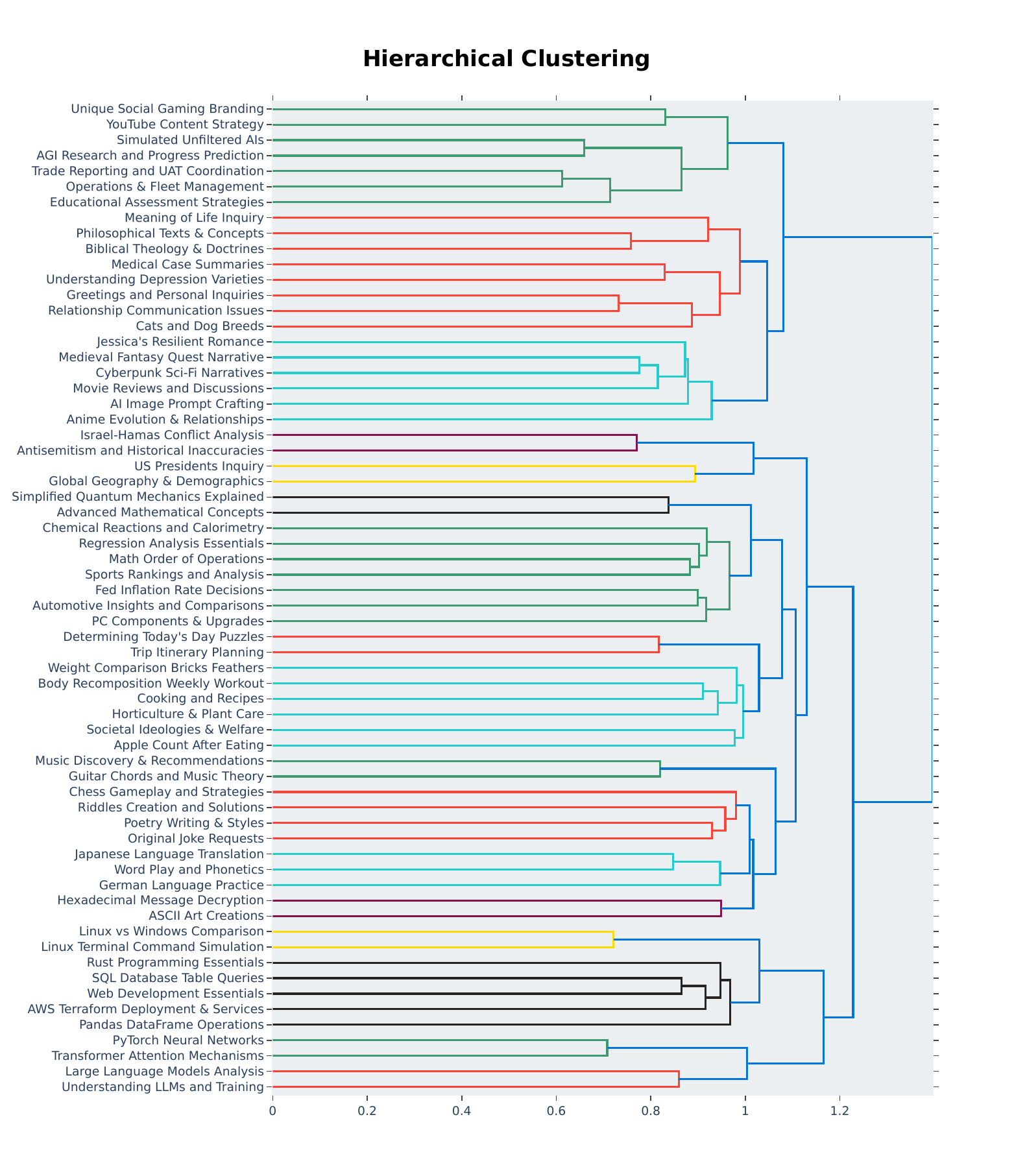}
\caption{Top-64 clusters visualized in hierarchy. x-axis represents the cosine similarity distance. y-axis shows the topic title per cluster summarized by gpt-4-turbo.}
\label{fig:hierarchy}
\end{figure}

\section{Confidence Interval Simulation Study}
\label{app:simulation-study}
\begin{figure*}[t]
\centering
\includegraphics[width=\textwidth]{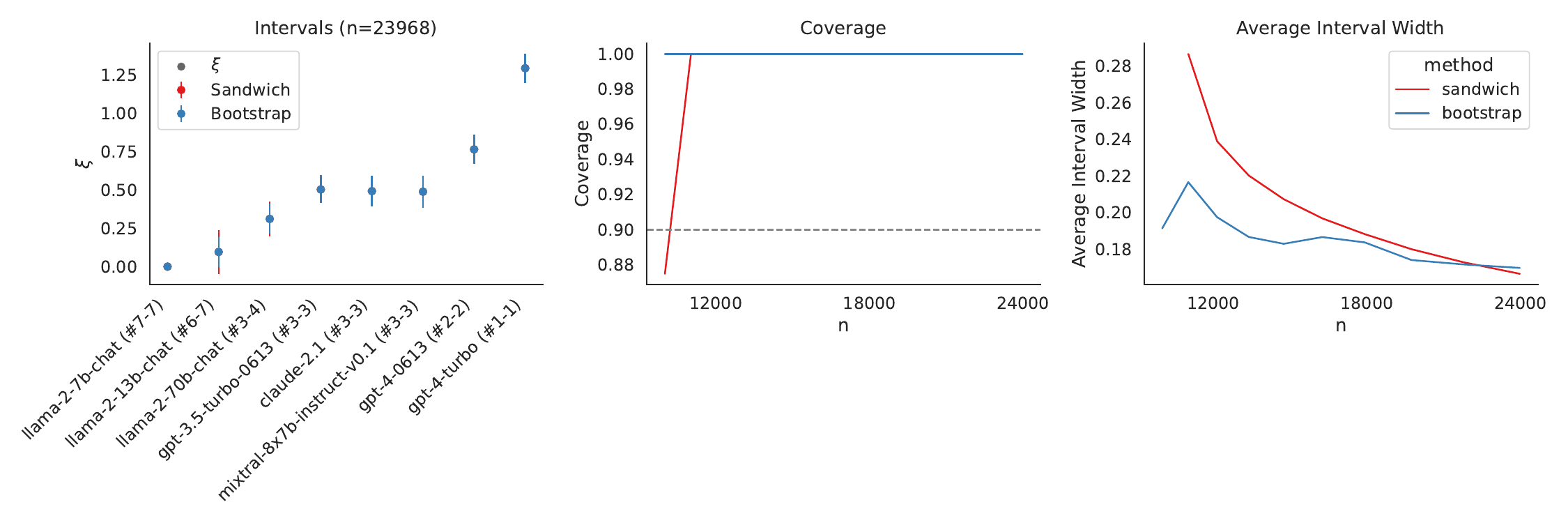}
\vspace{-20px}
\caption{Replay experiment showing the intervals, coverage, and average interval sizes of the bootstrap and of the sandwich intervals. The sandwich intervals, though larger in small samples, are more stable, and in large samples, they actually become smaller. We use the multiplicity corrected version of both intervals, so they both have a coverage of 1. (Coverage here is calculated with respect to the BT coefficient solution on the full dataset, so it is not as meaningful as in the simulation plot below.)}
\label{fig:bstrp-sndwch-rply}
\end{figure*}

We conduct a simulation study to evaluate the bootstrap confidence intervals versus the sandwich estimator.
To a large extent, both intervals are the same---indeed, their intervals are often identical to the naked eye.
Nonetheless, in our experiments, there are some differences.
First, in Figure~\ref{fig:bstrp-sndwch-rply}, we conduct a replay study using the same 213576 data points mentioned in the main text.

\begin{figure*}[t]
\centering
\includegraphics[width=\textwidth]{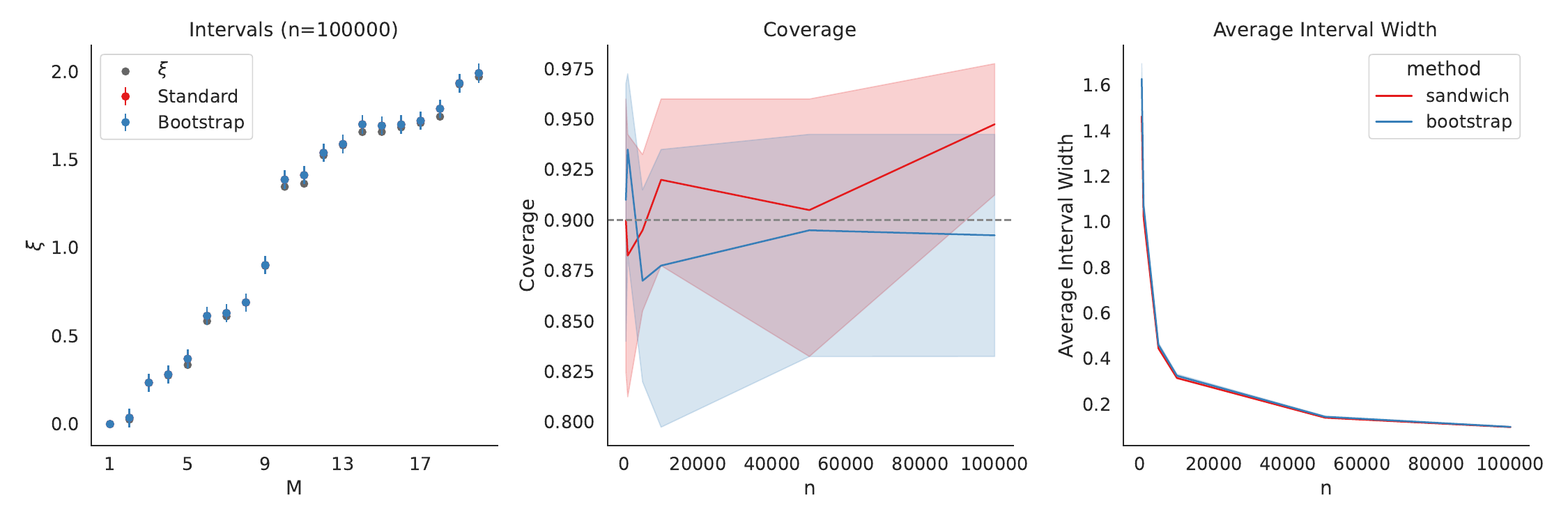}
\vspace{-20px}
\caption{Synthetic experiment. Coefficients are drawn from the BT-coefficient distribution $x$ on the left. Coverage of the uncorrected intervals is shown in the middle. Line plots of set width are shown on the right, and they almost perfectly match.}
\label{fig:btstrp-sndwich-synth}
\end{figure*}

We also do a suite of experiments in simulation using the same $\mathsf{beta}$ generating process as in the main text, with $\gamma=2$.
The result is shown in Figure~\ref{fig:btstrp-sndwich-synth}; results are similar across many choices of the parameter $\gamma$ and the model strength, which indicates that both intervals will have good coverage and width in the practical conditions we would expose them to.

\section{The Nonparametric Bradley-Terry Model}
\label{app:np-bt}

\textbf{Nonparametric Bradley-Terry.} We next consider a nonparametric extension of the Bradley-Terry (BT) model~\citep{bradley1952rank} to the case where the ranking is not necessarily transitive.
Let $\cG(m)$ denote the set of all \emph{paths} to the model $m$, i.e., 
\begin{equation}
    \cG(m) = \left\{g \in \cB^{M-1} : g_{i,1} \neq g_{j,1},\ \forall i\neq j,\ \text{ and } g_{M-1,2}=m \right\},
\end{equation}
where $\cB = \cA \cup \{ (a_2, a_1) : a \in \cA \}$.
Each element of $\cG(m)$ is a chain of model pairings that leads to $m$; for example, if $m=5$ and $M=6$, one element of $\cG(m)$ is $( (1, 2), (2, 4), (4, 3), (3, 6), (6, 5) )$.
Our score function is given by the average path-sum of the log odds of the second model winning, over the entirety of $\cG(m)$:
\begin{equation}
    \label{eq:bt-score}
    s(\theta)_m = \frac{1}{|\cG(m)|}\sum_{g \in \cG(m)} \left( \log \frac{\theta'((1,g_{1,1}))}{1-\theta'((1,g_{1,1}))} + \sum_{a \in g} \log \frac{\theta'(a)}{1-\theta'(a)}\right),
\end{equation}
where $\theta'(a) = \theta(a)\ind{a \in \cA} + (1-\theta((a_2, a_1)))\ind{a \notin \cA}$, with the convention that $\theta((m,m))=1/2$ for all $m$.
Note that for any $g \in \cG(m)$ where $a \in g$ and $m \notin a$, we also have some $g' \in \cG(m)$ such that $(a_2, a_1) \in g$.
Meanwhile, if $a \in g$ and $m \in a$, then $a = (m', m)$ for some $m'$.
Thus, we can compute
\begin{align}
    s(\theta)_m &= \sum\limits_{\substack{a \in \cA \\ m \notin a}} \frac{1}{2}\left(\log\frac{\theta'(a)}{1-\theta'(a)} + \log\frac{\theta'((a_2,a_1))}{1-\theta'((a_2,a_1))}\right) + \sum\limits_{m' \in [M]\setminus \{m\}} \left( \log\frac{\theta'((m',m))}{1-\theta'((m',m))} + \frac{\theta'((1,m'))}{1-\theta'((1,m'))} \right) \\
    & = \sum\limits_{\substack{a \in \cA \\ m \notin a}} \frac{1}{2}\left(\log\frac{\theta(a)}{1-\theta(a)} + \log\frac{1-\theta(a)}{\theta(a)}\right) + \sum\limits_{m' \in [M]\setminus \{m\}} \left( \log\frac{\theta'((m',m))}{1-\theta'((m',m))} + \frac{\theta'((1,m'))}{1-\theta'((1,m'))} \right) \\
    & = \sum\limits_{m' \in [M]\setminus \{m\}} \left( \log\frac{\theta'((m',m))}{1-\theta'((m',m))} + \frac{\theta'((1,m'))}{1-\theta'((1,m'))} \right) \\
    & = \sum\limits_{m' \in [M]\setminus \{m\}} \left( (1-2\ind{m' > m})\log\frac{\theta((m',m))}{1-\theta((m',m))} + \frac{\theta((1,m'))}{1-\theta((1,m'))} \right).
    \label{eq:np-bt-computable}
\end{align}

This score is always well-defined, and is a simple, smooth function of $\theta$.
Its derivative is, for all $a \in \cA$,
\begin{equation}
    \label{eq:bt-score-grad}
    \frac{\partial}{\partial \theta(a)} s(\theta)_m = \ind{a_2 = m}(1-2\ind{a_1 > m})\frac{1}{\theta(a)(1-\theta(a))} + \ind{a_1 = 1,\ a_2 \neq m}\frac{1}{\theta(a)(1-\theta(a))}.
\end{equation}

How is the BT score related to the original Bradley-Terry model?
In the original Bradley-Terry model, $H_t \in \{0, 1\}$, and the probability of model $m$ beating model $m'$ is assumed to be given by
\begin{equation}
    \label{eq:bt-prob}
    \theta((m',m)) = \frac{e^{\xi_m}}{e^{\xi_m} + e^{\xi_{m'}}}, 
\end{equation}
for some unknown parameters $\xi_1, \ldots, \xi_M$---the \emph{Bradley-Terry coefficients}.
The basic goal of the Bradley-Terry model is to estimate these parameters from the observed outcomes.
In our setting, however, we use the outcomes to get a CLT on $\theta$, and then can immediately recover the coefficients.
Taking without loss of generality $\xi_{1} = 0$, we have that
\begin{align}
    \log\frac{\theta((1,m'))}{1-\theta((1,m'))}+\log \frac{\theta((m', m))}{1-\theta((m', m))} &= \log\frac{\theta((1,m'))}{\theta((m',1))} + \log\frac{\theta((m',m))}{\theta((m,m'))}\\
    &= \log \frac{e^{\xi_{m'}}(e^{\xi_{m'}}+ 1)}{e^{\xi_{m'}}+1} + \log \frac{e^{\xi_m}(e^{\xi_{m'}}+ e^{\xi_m})}{e^{\xi_{m'}}(e^{\xi_{m'}}+ e^{\xi_m})} \\
    &= \xi_{m'} + \xi_m - \xi_{m'} = \xi_m
\end{align}
Thus, all the sums over paths in~\eqref{eq:bt-score} are equal to $\xi_m - \xi_{g_{1,1}}$.
\begin{align}
    &\log \frac{\theta'((1,g_{1,1}))}{1-\theta'((1,g_{1,1}))} + \sum\limits_{a \in g}\log \frac{\theta'(a)}{1-\theta'(a)} \\
    = & \xi_{g_{1,1}} + \xi_{g_{1,2}} - \xi_{g_{1,1}} + \xi_{g_{2,2}} - \xi_{g_{2,1}} + \cdots + \xi_{g_{M-1,2}} - \xi_{g_{M-1,1}} \\
    = & \xi_{g_{M-1,2}} = \xi_m.
\end{align}
Thus, if the parametric BT model is well-specified, the nonparametric version will exactly recover the Bradley-Terry coefficients.
However, our nonparametric analogue of the BT model has major advantages over the original: it will retain statistical validity even if $H_t$ is not binary, if the win rate is non-transitive, and if the logistic model assumed by the BT model is misspecified.
In practice, the nonparametric BT coefficient can be easily computed by~\eqref{eq:np-bt-computable}.

\section{Valid P-Value}
\label{app:p-val}
Consider the p-value
\begin{equation}
    p_i = \frac{1}{|\cH_{A'_i}|+1} \left(1 + \sum\limits_{h \in \cH_{A'_i}} \ind{h \geq H'_i}\right).
\end{equation}
We will prove that this p-value is valid, i.e., that $\P(p_i \leq t) \leq t$, under the null hypothesis that the vector $\cH' = (H_t : A_t = A'_i)\|(H'_i)$ is exchangeable, where $\|$ denotes the concatenation operator.
First, notice that we can equivalently write
\begin{equation}
    p_i = \frac{1}{|\cH'|} \sum\limits_{h \in \cH'} \ind{h \geq H'_i}.
\end{equation}
We also have that $\P(p_i \leq t) \leq \P(p_i \leq \frac{k}{|\cH'|})$, where $k = \lfloor t|\cH'| \rfloor /|\cH'|$.
Then, since the data points are exchangeable, we have that $H'_i$ is uniformly distributed among the ranks of $\cH'$, so $ \P(p_i \leq \frac{k}{|\cH'|}) \leq \frac{k}{|\cH'|} \leq t$, completing the proof.

\newpage
\lstset{
    basicstyle=\ttfamily,
    breaklines=true
}
\section{Sample Prompts}
\label{app:sample-prompts}

\color{black}\subsection{Prompt Examples From Different Clusters}
\label{subsec:clusters-example}
\textbf{Example 1:} An example user prompt from the cluster "Python Game Programming Challenge". This prompt distinctively separates GPT-4-0613 performance from Llama-2-70b-chat. The user request "higher quality code`` from each assistant. While GPT-4-Turbo provides a correct implementation along with helpful explanation, Llama-2-70b-chat fail to provide any program and provided a less helpful response overall. 
\begin{tcolorbox}
\texttt{\textbf{<|User Prompt|>}} \\
\texttt{Develop a snowman game similar to the Hangman game but draws a snowman figure standing on the snow instead of a hangman. It also downloads (or uses library) the complete English dictionary to select words for three levels of difficulty. Think step-by-step, take a deep breath and choose a higher quality code for each step.} \\

\texttt{\textbf{<|GPT-4-0613's Answer|>}} 
\begin{lstlisting}
Sure, here's a step-by-step guide on how to create a Snowman game similar to Hangman. We'll use Python for this example.
Step 1: Import necessary libraries
```python
import random
import urllib.request
```
Step 2: Download the English dictionary
```python
url = "https://raw.githubusercontent.com/dwyl/english-words/master/words.txt"
urllib.request.urlretrieve(url, 'words.txt')
```
Step 3: Load the words from the dictionary and categorize them into three levels of difficulty
```python
with open('words.txt') as word_file:
    english_words = word_file.read().split()
easy_words = [word for word in english_words if 4 <= len(word) <= 6]
medium_words = [word for word in english_words if 7 <= len(word) <= 9]
hard_words = [word for word in english_words if len(word) >= 10]
```
Step 4: Create a function to randomly select a word based on the difficulty level
```python
def get_random_word(difficulty):
    if difficulty == 'easy':
        word = random.choice(easy_words)
    elif difficulty == 'medium':
        word = random.choice(medium_words)
    else:
        word = random.choice(hard_words)
    return word.upper()
```
Step 5: Create a function to print the snowman
\end{lstlisting}
\end{tcolorbox}

\begin{tcolorbox}
\begin{lstlisting}
```python
def print_snowman(guesses):
    snowman = ['   _____   ', '  (     )  ', ' ( (   ) ) ', '(___(___)_)', '    ( )    ', '   (   )   ', '  (     )  ', ' (       ) ', '(         )', '  _______  ']
    for i in range(guesses):
        print(snowman[i])
```
Step 6: Create the main function for the game
```python
def play_snowman():
    difficulty = input("Choose difficulty (easy, medium, hard): ")
    word = get_random_word(difficulty)
    guessed_letters = []
    guesses = 0
    print('The word contains', len(word), 'letters.')

    while guesses < 10:
        guess = input("Guess a letter: ").upper()
        if guess in guessed_letters:
            print("You already guessed that letter.")
        elif guess in word:
            guessed_letters.append(guess)
            print("Good job! The word contains that letter.")
        else:
            guesses += 1
            print_snowman(guesses)
            print("Sorry, the word does not contain that letter.")
        print(' '.join(letter if letter in guessed_letters else '_' for letter in word))

        if set(guessed_letters) == set(word):
            print("Congratulations! You've successfully guessed the word.")
            break
    else:
        print("Sorry, you didn't guess the word. The word was", word)
```
Step 7: Run the game
```python
play_snowman()
```
This is a simple version of the game. You can enhance it by adding more features like scoring, time limit, etc.
```
\end{lstlisting}
\end{tcolorbox}

\begin{tcolorbox}
\texttt{\textbf{<|Llama-2-70b-chat's Answer|>}} 
    \begin{lstlisting}
Sure! Here's a step-by-step guide to developing a snowman game similar to Hangman, using the complete English dictionary for three levels of difficulty.

Step 1: Setting up the project

* Create a new Python project directory and initialize it with Git.
* Install the necessary libraries:
	+ `numpy` for generating random numbers
	+ `nltk` for accessing the English dictionary
	+ `pygame` for creating the game graphics
	+ `pydub` for playing sound effects (optional)
* Download the English dictionary dataset (e.g., from Kaggle) and store it in a file named `dictionary.txt`.

Step 2: Loading the dictionary and generating words

* Use the `nltk` library to read the dictionary file and store the words in a list.
* Split the list into three parts: easy, medium, and hard words, based on their length and frequency of use.
* Shuffle the lists to randomize the order of the words.

Step 3: Creating the game graphics

* Use `pygame` to create a window for the game, with a white background to represent the snow.
* Draw a snowman figure on the background, using black rectangles for the body, hat, and scarf.
* Add buttons for the player to submit their guesses.

Step 4: Implementing the game logic

* Create a function to generate a random word from the appropriate difficulty level and display it on the screen with missing letters (e.g., _ _ _ _ _ _).
* Create a function to check if a guessed letter appears in the word, and update the display accordingly (e.g., _ _ _ X _ _).
* Create a function to handle incorrect guesses, such as displaying a warning message or playing a sound effect.
* Create a function to win the game, such as displaying a congratulations message or playing a victory sound effect.

Step 5: Adding sound effects (optional)

* Use `pydub` to play sound effects for various actions in the game, such as guessing a letter correctly or incorrectly.
* Consider using a library like `pygame_mixer` to simplify the process of playing sound effects in pygame.
    \end{lstlisting}
\end{tcolorbox}

\begin{tcolorbox}
\begin{lstlisting}
Step 6: Testing and refining the game

* Test the game thoroughly to ensure that it works as expected, with no bugs or errors.
* Refine the game mechanics and graphics as needed, based on player feedback or personal preference.
* Consider adding additional features, such as keeping track of the player's score or implementing a time limit for each guess.

Step 7: Packaging and distributing the game

* Package the game as a standalone executable file using a tool like PyInstaller.
* Distribute the game to friends, family, or online communities, and gather feedback for future improvements.

This is a high-level overview of the steps involved in creating a snowman game similar to Hangman. Of course, the details will depend on your specific implementation and the tools you choose to use. Good luck with your project!
\end{lstlisting}

\texttt{\textbf{<|GPT-4-Turbo's Judgment Verdict|>}} \\
\texttt{GPT-4-0613 Wins}
\end{tcolorbox}

\textbf{Program Output:} Below is the terminal output of the snowman game programmed by GPT-4-0613, demonstrating its correctness. It is clear that GPT-4-0613's response is more helpful and relevant to the user's request.
\begin{tcolorbox}
\begin{lstlisting}
>>> python snowman.py
Choose difficulty (easy, medium, hard): easy
The word contains 6 letters.
Good job! The word contains that letter.
_ A _ _ _ _
Guess a letter: e
Good job! The word contains that letter.
_ A _ _ _ E
Guess a letter: d
   _____   
  (     )  
Sorry, the word does not contain that letter.
_ A _ _ _ E
Guess a letter: g
   _____   
  (     )  
 ( (   ) ) 
\end{lstlisting}
\end{tcolorbox}

\newpage
\textbf{Example 2:} An example user prompt from the cluster "Python Game Programming Challenge". This prompt distinctively separates GPT-4-0613 performance from Llama-2-70b-chat. Due to limited space, the code from each assistant's response is omitted.
\begin{tcolorbox}
\texttt{\textbf{<|User Prompt|>}} \\ 
\texttt{Write a python function that prints a christmas tree in the form of a triangle of asterisks, where the only input is size. }\\

\texttt{\textbf{<|GPT-4-0613's Code Output|>}} \\
\texttt{>{}>{}> print\_christmas\_tree(5)
    \begin{gather*}
    *    \\
    ***   \\
    *****  \\
    ******* \\
    *********
    \end{gather*}}

\texttt{\textbf{<|Llama-2-70b-chat's Code Output|>}} \\
\texttt{>{}>{}> print\_christmas\_tree(5) \\
* * * * * \\
* * * * \\
* * * \\
* * \\
* } \\

\texttt{\textbf{<|GPT-4-Turbo Verdict|>}} \\
\texttt{GPT-4-0613 Wins}
\end{tcolorbox}

\textbf{Example 3:} An example user prompt from the cluster "Movie Recommendations \& Rating". Both assistant retrieve the correct movie information and gave an almost identical explanation. Therefore, this prompt cannot distinguish GPT-4-0613's performance from Llama-2-70b-chat.
\begin{tcolorbox}
\texttt{\textbf{<|User Prompt|>} \\
Which film came out first, Blind Shaft or The Mask Of Fu Manchu?}\\

\texttt{\textbf{<|GPT-4-0613's Answer|>} \\
  The Mask of Fu Manchu was released in 1932, and Blind Shaft was released in 2003. Therefore, The Mask of Fu Manchu came out first.}\\

\texttt{\textbf{<|Llama-2-70b-chat's Answer|>}\\
The Mask Of Fu Manchu came out first. It was released in 1932, while Blind Shaft was released in 2003.} \\

\texttt{\textbf{<|GPT-4-Turbo Verdict|>} \\
Tie
}
\end{tcolorbox}

\newpage
\textbf{Example 4:} An example user prompt from the cluster "Travel Itinerary Planning". Both assistant provided an equally helpful itinerary. Therefore, this prompt cannot distinguish GPT-4-0613's performance from Llama-2-70b-chat. 
\begin{tcolorbox}
\texttt{\textbf{<|User Prompt|>} \\
Can you make a list of things I could do with my girlfriend in Lille, France?} \\

\texttt{\textbf{<|GPT-4-0613's Answer|>}}
\begin{lstlisting}
Absolutely, here are some suggestions for activities you and your girlfriend 
can enjoy in Lille, France:
1. Visit the Palais des Beaux-Arts: This is one of France's largest museums and it's right in the heart of Lille. It houses collections of European paintings, a medieval and Renaissance collection, ceramics, French sculptures and more.
2. Explore the Vieux Lille: Stroll through the charming old town with its Flemish influenced architecture, chic boutiques, and delightful eateries.
3. Visit the Lille Cathedral (Basilique-cathedrale Notre-Dame-de-la-Treille): This beautiful cathedral is a mix of modern and traditional architecture.
4. Enjoy a picnic at the Citadel Park: This is a great place to relax, have a picnic, and enjoy the outdoors.
5. Visit the LaM (Lille Metropole Musee d'art moderne, d'art contemporain et d'art brut): This museum is home to a unique collection of modern and contemporary art.
6. Take a boat tour on the Deule River: This is a romantic way to see the city and its surroundings.
7. Visit the Maison Natale de Charles de Gaulle: This is the birthplace of Charles de Gaulle and it's now a museum dedicated to his life and work.
8. Explore the Marche de Wazemmes: This is one of the largest markets in France and it's a great place to buy local products.
9. Enjoy a meal at a local restaurant: Lille is known for its delicious cuisine, so make sure to try some local dishes.
10. Visit the Lille Zoo: This is a small but well-kept zoo that houses about 70 different species.
11. Take a romantic walk through the Jardin des Plantes de Lille: This botanical garden is a peaceful place to enjoy nature.
12. Visit the Musee de l'Hospice Comtesse: This museum is located in a former hospital and it's home to a collection of art and artifacts from the Middle Ages.
13. Enjoy a performance at the Lille Opera: Check the schedule and book tickets for a romantic night out.
14. Take a day trip to the nearby city of Bruges: This charming Belgian city is just a short train ride away.
15. Visit the Le Palais Rihour: This palace houses the Lille Tourist Office and it's also home to the Salle du Conclave, which has a beautiful Gothic fireplace.
Remember to check the opening times and any COVID-19 restrictions before you 
visit. Enjoy your trip!
\end{lstlisting}
\texttt{\textbf{<|Llama-2-70b-chat's Answer|>}} \\
\texttt{Sure! Lille is a beautiful city in northern France that offers a variety of activities for couples to enjoy. Here are some ideas for things you could}
\end{tcolorbox}

\begin{tcolorbox}
\begin{lstlisting}
do with your girlfriend in Lille:
1. Explore the city's charming streets and historic landmarks, such as the Grand Place, the Vieille Bourse, and the Cathedral of Lille.
2. Visit one of the many museums, such as the Palais des Beaux-Arts de Lille, the Musee d'Art Moderne, d'Art Contemporain et d'Art Brut, or the Musee de l'Hospice Comtesse.
3. Take a stroll along the canal and enjoy the views of the city from the waterfront.
4. Visit the famous Lille Christmas market, which takes place every year in December and features festive stalls, food, and drinks.
5. Go shopping at the city's many boutiques and galleries, or at the large shopping centers such as the Euralille or the Lille Europe.
6. Enjoy a romantic dinner at one of Lille's many excellent restaurants, such as the Michelin-starred Le Comptoir du Relais or the cozy Le Bistrot d'Alsace.
7. Take a day trip to the nearby city of Bruges, Belgium, which is known for its picturesque canals and medieval architecture.
8. Go to a show or concert at the Lille Opera or the Le Zenith concert hall.
9. Visit the nearby town of Roubaix, which is famous for its textile industry and its beautiful Art Deco architecture.
10. Take a walk in the beautiful Parc de la Citadelle, which offers a peaceful escape from the city and features a pond, a botanical garden, and a zoo.
11. Go on a bike ride or a hike in the surrounding countryside, such as the Parc Naturel Regional Nord-Pas de Calais.
12. Visit the famous Lille flea market, which takes place every Sunday and offers a wide range of antiques, vintage items, and second-hand goods.
13. Take a cooking class together and learn how to make some of the region's delicious specialties, such as carbonnade flamande (beef stew) or tarte au maroilles (cheese tart).
14. Go on a wine tasting tour and discover the region's excellent wines, such as the Chablis or the Champagne.
15. Take a hot air balloon ride over the city and enjoy a bird's-eye view of Lille's beautiful architecture and surroundings.

These are just a few ideas to get you started. Lille is a vibrant and 
romantic city that offers many opportunities for couples to have a great 
time together. Bon voyage !
\end{lstlisting}

\texttt{\textbf{<|GPT-4-Turbo Verdict|>} \\
Tie}
\end{tcolorbox}

\newpage
\color{black}\subsection{Arena Bench Prompt}
\label{subsec:arena-bench-examples}

\textbf{Example 1}: An prompt from Arena Bench requiring a very high level of complex problem-solving skill and reasoning while adhering to real-world application.
\begin{tcolorbox}
    \texttt{Create a flutter app for habit tracking that tracks daily habits for a user. The user should be able create multiple daily tasks that he wants to track. And he should be able to group the habits into a set of groups. One the user completes a task, he is rewarded a number of points per task. The app should have a page that summarize the total score per group. And the aggregate score of all the groups of habits. This flutter app needs to be compilable for both andriod and iOS.}
\end{tcolorbox}

\textbf{Example 2}: An prompt from Arena Bench requiring a very high level of complex problem-solving skill and reasoning while adhering to real-world application.
\begin{tcolorbox}
    \texttt{I want to set up a remote raspberry pi zero, powered by a solar panel with simple wiring. I want to power a small 2W pump, a simple electet microphone, a custom python script running on the raspberry pi that is used to classify audio detected by the microphone. What components will I need to optimise for cost and minimise any electrical work (e.g. soldering)? What size solar panel will I need to power this whole system?}
\end{tcolorbox}

\newpage
\color{black}\subsection{Arena Bench System Prompt}
\label{subsec:arena-bench-sys-prompt}
The novel evaluation procedure is as follow: we prompt GPT-4-Turbo with the system prompt displayed below alongside a user prompt, a reference answer, and 2 assistant's answers. For reference answer, we present the user prompt with 3 assistants' answers, GPT-4-Turbo, GPT-4-0314, and Claude-1, to GPT-4-Turbo and ask GPT-4-Turbo to generate an answer to the prompt. To ensure consistent pairwise judgment, we set up GPT-3.5-Turbo-0301 as the baseline answer for all models to be compared against. To avoid positional bias, we conduct two judgments per prompt: the first judgment presents the baseline answer as Assistant A while the second judgment presents the baseline answer as Assistant B. In total, we conduct 700 pairwise comparisons between each model against GPT-3.5-Turbo-0301 across 350 user prompts to calculate a win-rate against the baseline. Then we project the win-rate on a scale from 0 to 10 by assigning wins with a score of 10, ties with a score of 5, and losses with a score of 0. Further, we assign a significant win or loss as 3 wins or 3 losses, respectively, and keeping the other verdicts as a single win, loss, or tie. Finally, we calculate the final score by averaging across the wins, losses, and ties.

\begin{tcolorbox}
\texttt{\textbf{<|System Prompt|>}} \\
\texttt{Please act as an impartial judge and evaluate the quality of the responses provided by two AI assistants to the user prompt displayed below. Your job is to evaluate which assistant's answer is better.}\\

\texttt{When evaluating the assistants' answers, compare both assistants' answers. You must identify and correct any mistakes or inaccurate information.}\\

\texttt{Then consider if the assistant's answers are helpful, relevant, and concise. Helpful means the answer correctly responds to the prompt or follows the instructions. Note when user prompt has any ambiguity or more than one interpretation, it is more helpful and appropriate to ask for clarifications or more information from the user than providing an answer based on assumptions. Relevant means all parts of the response closely connect or are appropriate to what is being asked. Concise means the response is clear and not verbose or excessive.}\\

\texttt{Then consider the creativity and novelty of the assistant's answers when needed. Finally, identify any missing important information in the assistants' answers that would be beneficial to include when responding to the user prompt.}\\

\texttt{After providing your explanation, you must output only one of the following choices as your final verdict with a label:}\\

\texttt{1. Assistant A is significantly better: [[A>{}>B]]\\
2. Assistant A is slightly better: [[A>B]]\\
3. Tie, relatively the same: [[A=B]]\\
4. Assistant B is slightly better: [[B>A]]\\
5. Assistant B is significantly better: [[B>{}>A]]} \\

\texttt{Example output: "My final verdict is tie: [[A=B]]".} \\




\end{tcolorbox}

\newpage
\color{black}\subsection{Arena Battles with No Ground Truth}
\label{subsec:no-ground-truth-example}

\textbf{Example 1}: An example of an Arena Battle between GPT-4-Turbo and Llama-2-13B-chat, where depending on different preferences, either answer can be argued as better than the other. While both answers are correct, GPT-4-Turbo's answer is concise and appropriate to the format of multiple choice question. On the other hand, Llama-2-13B-chat provides helpful explanation for the multiple choice answer.
\begin{tcolorbox}
\texttt{\textbf{<|User Prompt|>}} \\
\texttt{Multiple Choice Question \\
All plants have both a haploid and a diploid form. As the plant changes from one form to another, this is called:} \\

\texttt{Multiple choice question.\\
alternative life cycle\\
succession\\
an alternation of generations} \\

\texttt{\textbf{<|GPT-4-Turbo's Answer|>} \\
an alternation of generations}\\

\texttt{\textbf{<|Llama-2-13B-chat's Answer|>} \\
The correct answer is: an alternation of generations. }\\

\texttt{Plants, like other organisms, have different life cycle stages, and they can exist in two different forms: haploid and diploid. The haploid form is a single set of chromosomes, while the diploid form is two sets of chromosomes. As the plant changes from one form to another, this is called an alternation of generations. This process involves the alternation of haploid and diploid phases, and it is a fundamental aspect of the life cycle of plants.}\\

\texttt{Alternative life cycle and succession are not correct descriptions of this process, as they refer to different concepts. Alternative life cycle refers to the different stages of life an organism may go through, such as larva, pupa, and adult in insects, while succession refers to the process of one species replacing another in a particular environment or niche.}
\end{tcolorbox}

\textbf{Example 2}: An example of an Arena Battle between GPT-4-Turbo and Llama-2-13B-chat, where depending on different preferences, either answer can be argued as better than the other.
\begin{tcolorbox}
\texttt{\textbf{<|User Prompt|>} \\
What are critical elements of a Product Mindset for startup founders?} \\

\texttt{\textbf{<|GPT-4-Turbo's Answer|>} \\
Adopting a Product Mindset is crucial for startup founders as it helps them focus on creating products that deliver value to customers while achieving business objectives. Here are some critical elements of a Product Mindset for startup founders:} \\

\texttt{1. Customer Centricity: \\
   - Understand and empathize with your customers’ needs, problems, and motivations.}
\end{tcolorbox}

\begin{tcolorbox}
\texttt{- Continuously gather and incorporate customer feedback into the product development process. \\
   - Prioritize features and improvements that directly address customer pain points and enhance their experience.} \\

\texttt{2. Value Proposition: \\
   - Clearly define the unique value your product offers to the market. \\
   - Align your product's features and capabilities with its value proposition to ensure it meets the expectations of your target audience.} \\

\texttt{3. Problem-Solving Focus: \\
   - Identify and focus on solving real problems for users rather than just creating features for the sake of innovation. \\
   - Validate assumptions and hypotheses through experimentation and user testing.} \\

\texttt{4. Iterative Development: \\
   - Embrace an agile development approach, where you build, measure, and learn in iterative cycles. \\
   - Be open to pivoting or making significant changes based on feedback and market response.} \\

\texttt{5. Data-Driven Decision Making: \\
   - Use data to inform decisions about product development, marketing, and growth strategies. \\
   - Establish key performance indicators (KPIs) to measure the success of the product and track progress over time.} \\

... \\

\texttt{By internalizing these elements, startup founders can develop a strong Product Mindset that guides their decisions and strategies, ultimately leading to a product that customers love and that contributes to the startup's success.} \\

\texttt{\textbf{<|Llama-2-13B-chat's Answer|>} \\
A product mindset is a way of thinking and approaching product development that is critical for startup founders. Here are some essential elements of a product mindset for startup founders:} \\

\texttt{1. Customer-centricity: A product mindset starts with understanding the customer's needs, desires, and pain points. Founders must be obsessed with their customers and be willing to put in the effort to understand their needs. \\
2. Iteration and experimentation: A product mindset is all about experimentation and iteration. Founders must be willing to try new things, test hypotheses, and iterate on their product based on customer feedback.} \\
\texttt{3. Focus on value: A product mindset is focused on delivering value to the customer. Founders must be able to identify and prioritize the most important features and functionality that deliver the most value to their customers.} \\
\end{tcolorbox}

\begin{tcolorbox}
\texttt{4. Collaboration: A product mindset requires collaboration between cross-functional teams, including engineering, design, and product management. Founders must be able to work effectively with these teams to develop a product that meets customer needs.}\\
\texttt{5. Data-driven decision making: A product mindset is data-driven, and founders must be comfortable using data to make informed decisions about their product. This includes analyzing customer feedback, usage data, and other metrics to inform product development.}\\

...\\

\texttt{By embracing these elements of a product mindset, startup founders can develop a product that meets customer needs, delivers value, and sets their company up for long-term success.}
\end{tcolorbox}


\end{document}